\documentclass[10pt,twocolumn,letterpaper, hyphens, dvipsnames, table]{article}

\usepackage{listings}
\usepackage{multirow}
\usepackage{colortbl}
\usepackage{array} 
\usepackage{float}
\definecolor{mypink}{RGB}{255,200,190} 
\definecolor{myblue}{RGB}{203,203,254} 
\usepackage{capt-of} 
\usepackage{graphicx}
\usepackage{times}

\usepackage[pagenumbers]{iccv} 
\definecolor{iccvblue}{rgb}{0.21,0.49,0.74}
\usepackage[pagebackref,breaklinks,colorlinks,allcolors=iccvblue]{hyperref}
\usepackage[accsupp]{axessibility}  
\usepackage{xurl} 
\title{DualReal: Adaptive Joint Training for Lossless Identity-Motion Fusion 

in Video Customization}

\author{Wenchuan Wang, Mengqi Huang, Yijing Tu, Zhendong Mao\footnotemark[1]\\
University of Science and Technology of China\\
{\tt\small \{wenc\_k, huangmq, tuyijing\}@mail.ustc.edu.cn, zdmao@ustc.edu.cn}
}

\begin{document}
\twocolumn[{
    \renewcommand\twocolumn[1][]{#1}
    \maketitle
    \vspace{-1.35em}
    \begin{center}
        \includegraphics[width=1\linewidth]{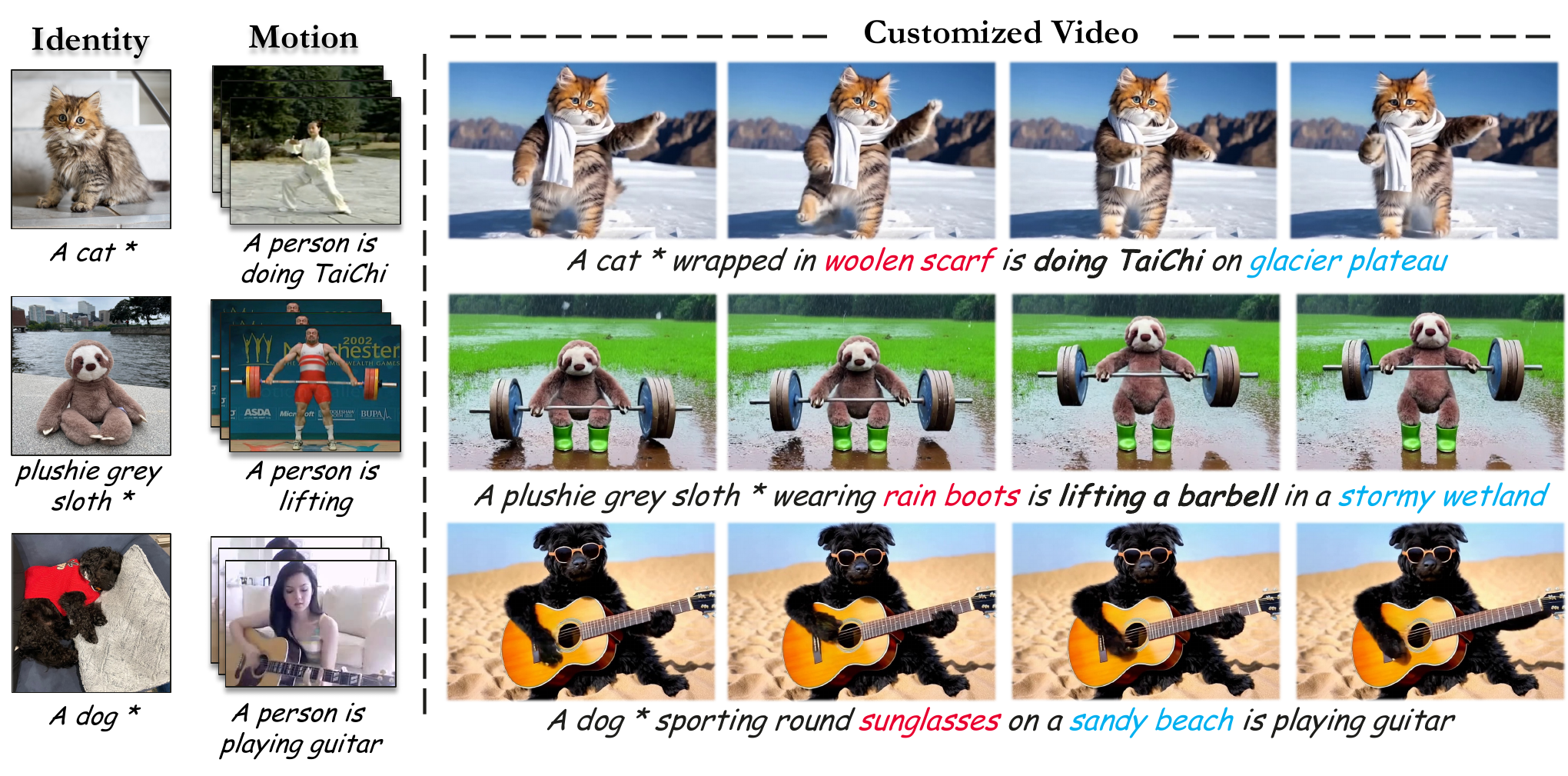}
        \vspace{-1em}
        \captionof{figure}{Generated customization results of our proposed novel paradigm \textbf{\textit{DualReal}}. Given identity images and motion videos, \textit{DualReal} generates high-quality customized identity and motion simultaneously, without compromising the consistency of either dimension.}
        \label{fig:result}
    \end{center}
    \vspace{-0.1em}
}]
\renewcommand{\thefootnote}{\fnsymbol{footnote}}
\footnotetext[1]{Corresponding author}

\begin{abstract}
Customized text-to-video generation with pre-trained large-scale models has recently garnered significant attention by focusing on identity and motion consistency.
Existing works typically follow the isolated customized paradigm, where the subject identity or motion dynamics are customized exclusively.
However, this paradigm completely ignores the intrinsic \textbf{mutual constraints and synergistic interdependencies} between identity and motion, resulting in identity-motion conflicts throughout the generation process that systematically degrade.
To address this, we introduce \textbf{DualReal}, a novel framework that employs adaptive joint training to construct interdependencies between dimensions collaboratively. Specifically, DualReal is composed of two units: (1) \textbf{Dual-aware Adaptation} dynamically switches the training step (\ie, identity or motion), learns the current information guided by the frozen dimension prior, and employs a regularization strategy to avoid knowledge leakage; (2) \textbf{StageBlender Controller} leverages the denoising stages and Diffusion Transformer depths to guide different dimensions with adaptive granularity, avoiding conflicts at various stages and ultimately achieving lossless fusion of identity and motion patterns. We constructed a more comprehensive evaluation benchmark than existing methods. The experimental results show that \textit{DualReal} improves CLIP-I and DINO-I metrics by \textbf{21.7\%} and \textbf{31.8\%} on average, and achieves top performance on nearly all motion metrics. Page: \url{https://wenc-k.github.io/dualreal-customization}
\end{abstract}
    
\section{Introduction} 
Video constitutes a spatiotemporal embodiment of the real world, where the spatial \emph{subject identity} and the temporal \emph{motion dynamics} form \emph{\textbf{mutually constrained yet synergistic dimensions}} for physical modeling. 
This mutuality manifests through their inherent interdependence, \ie, maintaining stable subject identities across frames restricts motion possibilities, while enforcing certain motion trajectories conversely necessitates corresponding topological transformations of identity representation (\eg, 180° viewpoint rotation leads to identity transformation from frontal to dorsal profiles).

Video customized generation \cite{jiang2024videobooth, wang2024customvideo, zhao2024motiondirector, wu2024motionbooth, wei2024dreamvideo}, which aims to mimic the user-specified concepts (\ie, subject identities, dynamic motions, or both) beyond linguistic expressibility, significantly enhances the controllability of video synthesis systems. 
This task greatly expands the applicability scope of pre-trained text-conditioned video models to cinematic production, personalized avatars, \etc, attracting growing interest from academic and industrial communities. 
The primary challenge of customized video generation lies in two interdependent dimensional objectives, \ie, 
(1) \textbf{\emph{identity consistency}}, \ie, the target subject should closely match the given reference in all frames, while \emph{minimizing temporal motion artifacts}, 
and (2) \textbf{\emph{motion consistency}}, \ie, the subject motion should closely match the given reference across frames, while \emph{minimizing spatial identity artifacts}.


As a rapidly emerging research frontier, existing video customized methodologies currently focus on either identity or motion independently. 
VideoBooth \cite{jiang2024videobooth} achieves subject identity-driven generation by injection of reference image embedding, while AnimateDiff \cite{guo2023animatediff} achieves the animation of static outputs into videos by appending trainable temporal modules to personalized text-to-image models. 
Recently, DreamVideo \cite{wei2024dreamvideo} employs independent training for each dimension (\ie, identity or motion) and directly blends their parameters during inference to achieve both the identity and motion customization simultaneously, demonstrating promising results to combine specific subject identity and motion patterns. 
Essentially, current works typically follow the isolated customized paradigm, where the subject identity or motion dynamics are customized exclusively.

\begin{figure}[t]
    \centering
    \includegraphics[width=1\linewidth]{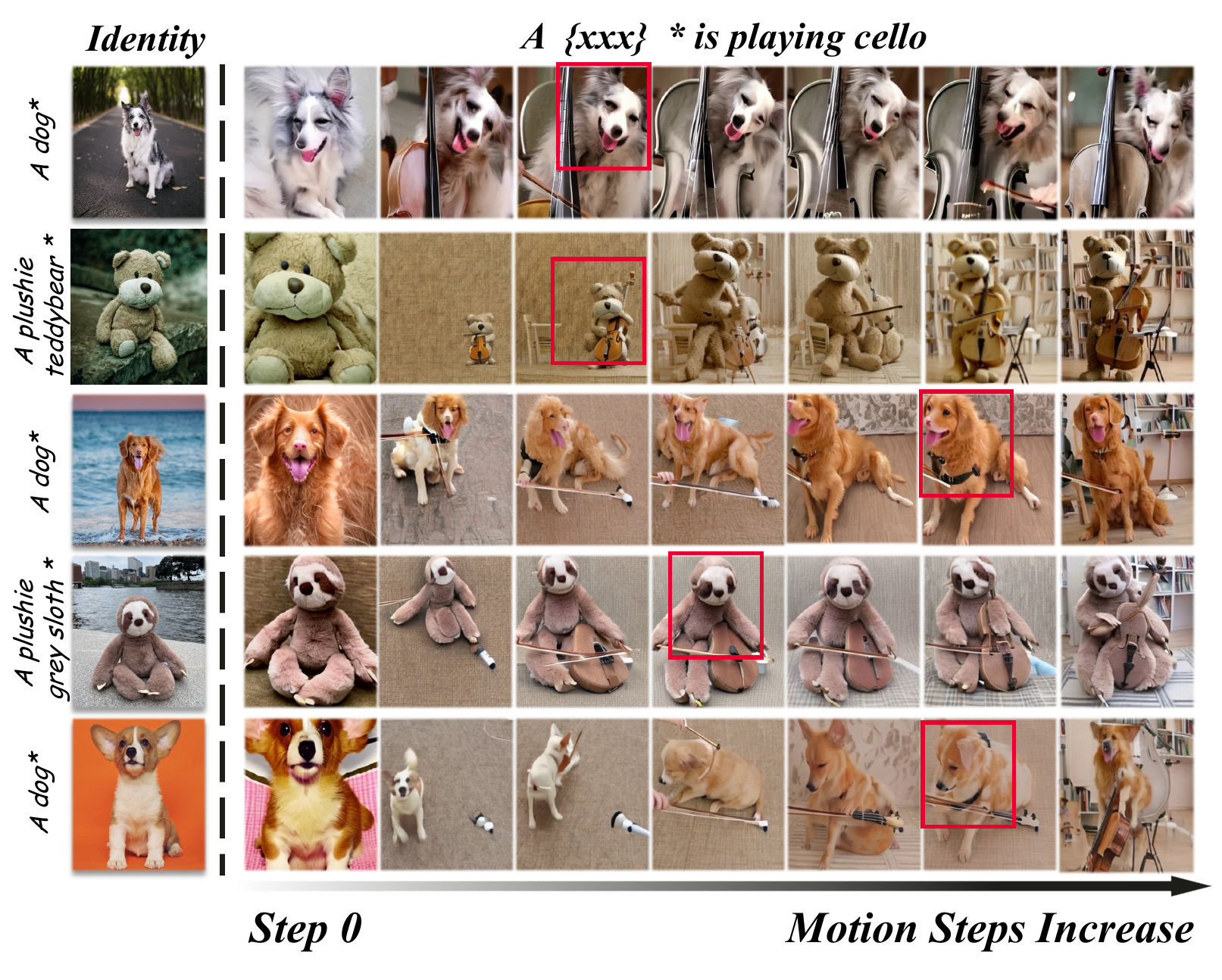}
    \caption{
    \textbf{Visual analysis of isolated training paradigm.} We select different identities with the same motion pattern, fix the number of identity training steps, and gradually increase the number of motion training steps to achieve two-dimensional customization. \emph{The red box marking the relative optimal position of the same identity's fidelity.} Experiments show that (1) \emph{adding motion prior significantly damages identity consistency}; (2) \emph{We cannot find a universal step to minimize identity degradation} from the positions of the red boxes for different identities.
    }
    \label{fig:motivation}
\end{figure}

However, the existing isolated customized paradigm completely ignores the intrinsic \emph{mutual constraints and synergistic interdependencies} between identity and motion, resulting in identity-motion conflicts throughout the generation process that systematically degrade either motion coherence, subject fidelity, or both dimensions simultaneously. 
The reason is that the step-by-step diffusion video synthesis process itself, by nature, with different denoising steps dynamically reweights their spatiotemporal focus, \ie, progressively refines identity details across frames, enabling complete modeling through increasing denoising steps.
The existing isolated customized paradigm violates this natural progression through indiscriminate dimensional over-specialization across all time steps, since they enforce uniform step sampling during motion/identity customized training, thereby inducing conflicting optimization trajectories between motion and identity accuracy. 
Consequently, existing methodologies inevitably cause mutual performance deterioration, that is, motion customization undermines the pre-trained video model's inherent identity priors, or \emph{vice versa}.
As shown in~\cref{fig:motivation}, (1) adding the motion prior \emph{irreversibly reduced identity fidelity}, indicating that isolated training fails to resolve dimensional conflicts during inference and leads to performance degradation; and (2) as motion training steps increased, optimal fidelity for different identities occurred unpredictably, suggesting that \emph{no universal number of training steps minimizes degradation}. In summary, the existing paradigm does not meet the consistency and flexibility requirements of video customization tasks.

In this paper, we introduce \textbf{\textit{DualReal}}, a novel framework that, for the first time, employs adaptive joint training to collaboratively construct interdependencies between identity and motion, which meets the consistency requirements of customized video generation in both identity and motion, as shown in \cref{fig:result}.
Technically, \textit{DualReal} is composed of two complementary units: (1) 
\textbf{\textit{Dual-aware Adaptation}} dynamically switches the training step (\ie, either identity or motion) to learn the current information, guided by the other dimension prior. It also employs a regularization strategy to prevent dimensional knowledge leakage by blocking the parameter updates of the non-training dimension using gradient masking, thereby achieving effective joint training; (2) \textbf{\textit{StageBlender Controller}} operates through coordinated utilization of denoising-stage progression and Diffusion Transformer (DiT) layer-depth variations during training. By adaptively allocating hierarchical focus (\ie, fine-grained adjustments for identity patterns and motion dynamics), it resolves dimensional competition across processing stages. This granularity-aware guidance ultimately achieves dimensional lossless fusion.

Our contributions can be summarized as follows:

\textcolor{blue}{Concepts.} For the first time, we (1) point out that the isolated paradigm causes mutual performance deterioration (identity fidelity and motion coherence) because it ignores the intrinsic constraints and synergistic; (2) present \textit{DualReal}, a novel paradigm that employs adaptive joint training to collaboratively construct interdependencies.

\textcolor{blue}{Technology.} The proposed \textit{DualReal} framework consists of two components: (1) \textit{Dual-aware Adaptation}, which alternates between identity and motion training phases, leveraging dimension-specific guidance and regularization to prevent information leakage and enable joint training; (2) \textit{StageBlender Controller}, which adaptively coordinates denoising stages and DiT depths to guide modes at different granularities, resolving conflicts and enabling seamless fusion of identity and motion patterns.

\textcolor{blue}{Performance} We constructed a more comprehensive evaluation benchmark than existing methods. The experimental results show that \textit{DualReal} improves CLIP-I and DINO-I metrics by \textbf{21.7\%} and \textbf{31.8\%} on average, and achieves top performance on nearly all motion quality metrics, demonstrating the efficiency of our framework.
\section{Related Work}
\subsection{Text-to-video Diffusion Models}
Recent advances in generative models have significantly improved the quality and versatility of synthetic content\cite{huang2023towards, huang2022dse, ho2022video, singer2022make, zhou2022magicvideo, esser2023structure}. 
DQVAE~\cite{huang2023towards} generates images autoregressively in a more effective coarse-to-fine order.
Text-to-video generation aims to generate realistic videos based on prompts and has recently received growing attention
\cite{an2023latent, zhang2024show, qing2024hierarchical, wang2023modelscope, guo2023animatediff, wu2023tune, yang2024cogvideox, wang2023videocomposer, tan2024animate, sora2024}.
Current text-to-video generation architectures primarily fall into two categories \cite{ronneberger2015u, peebles2023scalable}.
UNet-based video diffusion frameworks utilize hierarchical enc-dec with spatiotemporal learning
\cite{singer2022make, guo2023animatediff, hong2022cogvideo, blattmann2023stable, wu2023tune, wang2023modelscope, ho2022video},
\eg, Video diffusion models~\cite{ho2022video} pioneered diffusion model applications in video generation through pixel-space video distribution modeling.
Make-A-Video~\cite{singer2022make} and AnimateDiff~\cite{guo2023animatediff} augment pretrained text-to-image models with motion modules.
Recent advances in scalability drive the shift toward transformer-based architectures with joint spatiotemporal modeling \cite{sora2024,yang2024cogvideox, jabri2022scalable, menapace2024snap, kong2024hunyuanvideo}, achieving revolutionary progress in video generation. Sora~\cite{sora2024} introduces the diffusion-transformer framework, achieving cinematic-quality extended video synthesis with temporal stability. CogVideoX~\cite{yang2024cogvideox} introduces an expert transformer for enhanced text-video feature fusion.
While diffusion transformers exhibit strong generative capacities, their architectural constraints in spatiotemporal decoupling inherently limit dynamic concept embedding.

\subsection{Generation Model Customization}
Generation Model Customization has emerged as a pivotal strategy\cite{shi2024instantbooth,smith2023continual, chen2025customcontrast, huang2024realcustom, mao2024realcustom++}. In contrast to domain-agnostic generation frameworks, customized visual synthesis exhibits superior adaptability in addressing personalized visual requirements via parametric adaptation mechanisms\cite{chen2023disenbooth, chen2023subject, han2023svdiff, ruiz2023dreambooth, ruiz2024hyperdreambooth, wei2023elite, gal2022image, huang2023towards}.
Textual inversion~\cite{gal2022image} aligns visual-textual semantics through text embedding optimization.
Dreambooth~\cite{ruiz2023dreambooth} through full model fine-tuning of diffusion architectures to inject subject-specific priors.
RealCustom\cite{huang2024realcustom, mao2024realcustom++} disentangles similarity from controllability by precisely limiting subject influence to relevant parts only.
Building upon these foundational approaches, contemporary video customization research explores analogous methodologies\cite{molad2023dreamix, chefer2024still, ma2024magic, he2024id, wei2024dreamvideo, wu2024motionbooth, ren2024customize, bai2024uniedit, geyer2023tokenflow}.
MotionBooth~\cite{wu2024motionbooth} proposes a comprehensive video diffusion model fine-tuning coupled with attention map manipulation for motion control during inference.
DreamVideo~\cite{wei2024dreamvideo} develops decoupled adapter training with joint inference mechanisms, coordinating subject customization and motion preservation during generation.
\section{Methodology}
\begin{figure*}[t]
    \centering
    \includegraphics[width=\linewidth]{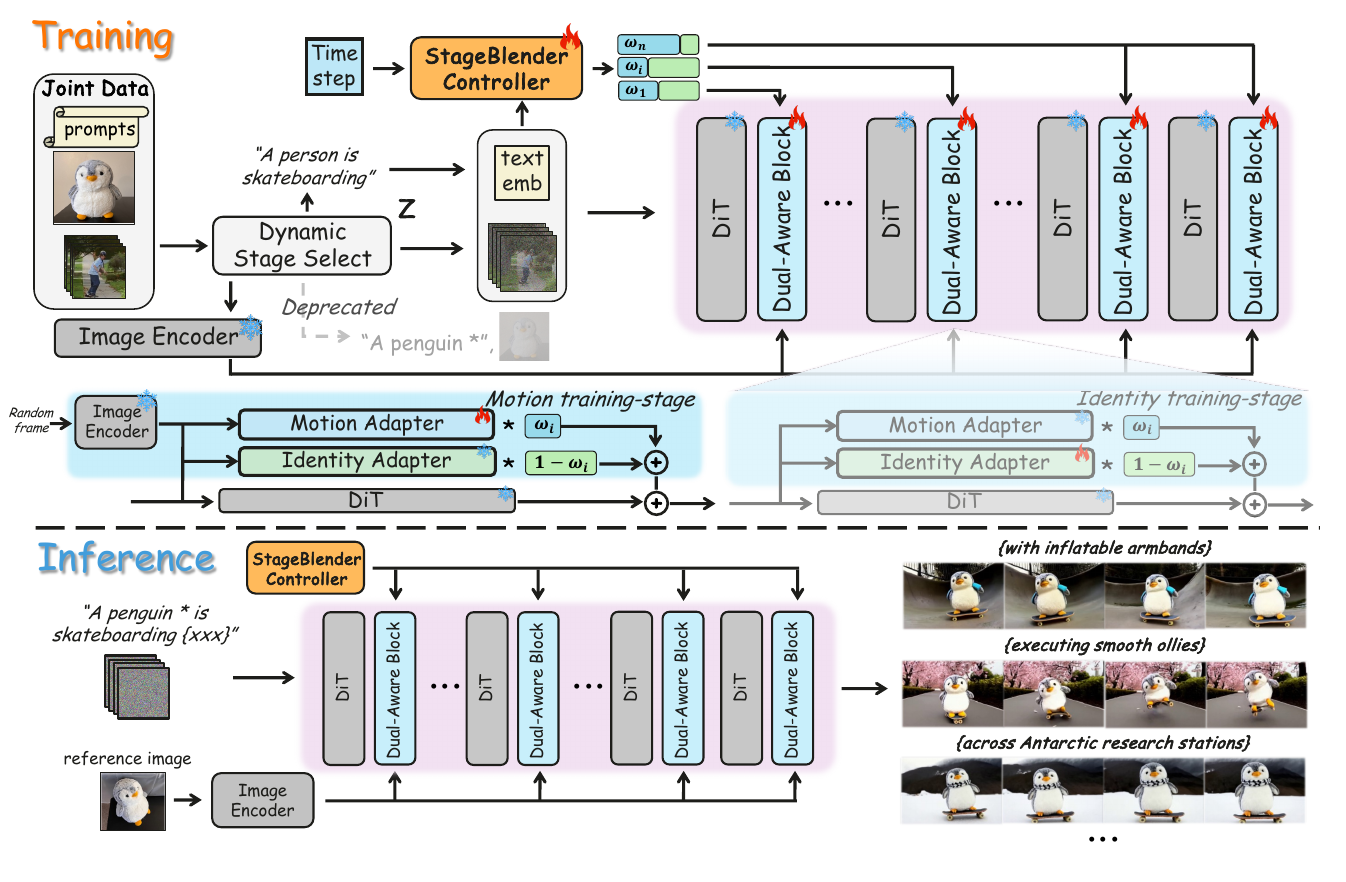}
    \caption{\textbf{Overall framework of DualReal}. At each training step, we first dynamically switch the training step $Z$ (\ie, identity or motion) to determine the data processing path. The specific data undergoes noise injection and combines with the text embeddings. \textit{StageBlender Controller} governs two-dimensional adapters' contributions in \textit{Dual-Aware Block} (DA-Block) through time-aware conditioning of current denoising step and fused feature representations. 
    In DA-Block, the training-stage adapter learns the current information guided by the frozen dimension prior, and employs a regularization strategy to avoid dimensional knowledge leakage, achieving joint training. Both branches engage in residual connections with DiT outputs.}
    \label{fig:pipeline}
\end{figure*}
Given a series of identity-specific images and motion sequences, \textit{DualReal} through an innovative joint training framework, synthesizes coherent motion while preserving full-frame identity fidelity, as detailed in \cref{fig:pipeline}.
During training, \textit{DualReal} dynamically switches optimization focus between identity and motion,  adjusting the module parameters through specific guidance.
To address the key challenges: (1) enabling joint identity-motion interaction modeling in unified parameter spaces. (2) mitigating attribute leakage risks in alternating training dimensions. We propose the \textbf{Dual-aware Adaptation} architecture with a complementary regularization strategy in \cref{sec:dual-aware}, forming an integrated framework for parameter-shared adaptation.
Moreover, we propose \textbf{StageBlender Controller}, which leverages the denoising stages and DiT depths to guide different dimensions with adaptive granularity, avoiding conflicts at various stages and ultimately achieving high-fidelity fusion of identity and motion in \cref{sec:hypernet}.


\subsection{Preliminary}
\label{sec:preliminary}

\noindent\textbf{DiT-based Models.}  Most DiT-based models \cite{yang2024cogvideox, sora2024, kong2024hunyuanvideo, menapace2024snap} process concatenated conditioning prompt and spatiotemporal visual tokens through transformer layers, establishing multi-modal coupling between text-guided semantic contexts and visual representations in latent space. Despite achieving remarkable capabilities in generic generation scenarios, this architecture fundamentally constrains conventional personalization frameworks that demand decoupled control along spatial-temporal axes \cite{wang2024customvideo, wu2024motionbooth}.
\subsection{Dual-aware Adaptation}
\label{sec:dual-aware}
To achieve joint training of identity and motion while resolving dimensional conflicts, we innovatively proposed \textit{Dual-aware Adaptation}, which leverages the prior from one dimension to guide the training of the other while preventing information leakage through a regularization strategy, as shown in the lower half of \cref{fig:pipeline}.

\noindent\textbf{Joint Identity-Motion Optimization.}
Different from some approaches that fine-tune the whole diffusion model ~\cite{ruiz2023dreambooth}, \textit{DualReal} first dynamically switches the training step (\ie, motion-focused or identity-focused) with the predefined hyperparameter ratio before each denoising iteration; the corresponding data is then sent to the DiT.
The input of $i$-th block is the joint feature $f_{\text{in}}^i=[f_\text{text}^i,f_\text{visual}^i] \in \mathbb{R}^{B \times (n_t+n_v) \times c}$, where $n_t$, $n_v$ represent the number of text and visual tokens respectively. The adapters employ the bottleneck architecture with skip connections ~\cite{wei2024dreamvideo}:
\begin{align}
&  f_{\text{id}}^i = \sigma\left(f_{\text{in}}^i * \mathbf{W}_{\text{down}} * \mathbf{W}_{\text{up}}\right), \\
&  f_{\text{mo}}^i = \sigma\left((f_{\text{in}}^i*\mathbf{W}_{\text{cond}}) * \mathbf{W}_{\text{down}}^{'} * \mathbf{W}_{\text{up}}^{'} \right),
\end{align}
where the activation function $\sigma$ corresponds to GELU ~\cite{hendrycks2016gaussian}, $\mathbf{W}$ and $\mathbf{W^{'}}$denote the identity and motion linear projection weights, respectively, both operating on hidden dimension $d$. The weight $\mathbf{W}_\text{{cond}}\in \mathbb{R}^{e \times c}$ of conditional linear maps reference image embedding to the latent space~\cite{wei2024dreamvideo}.

Through StageBlender Controller constraint (\cref{sec:hypernet}), the motion adapter outputs are scaled by weight coefficient $\omega_i$, with identity outputs weighted by the complementary coefficient ($1-\omega_i$). The modulated features are aggregated into the output of DiT blocks through residual connections. The above process can be formulated as:
\begin{align}
    \hat{f}_{\text{out}}^i = \omega_{i}*f_{\text{mo}}^i + (1-\omega_i)*f_{\text{id}}^i+f_\text{dit}^i\ ,
\end{align}
where $f_\text{dit}^i$ denotes the output of the $i$-th DiT layer and $\hat{f}_{\text{out}}^i$ indicates the aggregated output of the final block. This parametric constraint intrinsically balances feature contributions across blocks and denoising stages, while structurally enforcing dedicated attention to either identity preservation or motion dynamics during adaptation.

\noindent\textbf{Regularization Strategy.} A critical challenge in joint dimensional training arises from the significant distribution shift across different training dimensions, where unconstrained optimization usually causes destructive interference between cross-dimension knowledge observed in previous work~\cite{wu2024customcrafter, wu2024motionbooth}. For example, fine-tuning the motion adapter with static images during the temporal training phase irreversibly degrades its dynamic generation capability, with analogous effects occurring during identity adaptation.
To resolve this, we employ regularization with the gradient mask $M$ to activate only the corresponding adapter parameters based on a binary selector variable $Z \in \{0,1\}$, optimizing motion coherence(\ie, through motion adapter), or preserving identity consistency(\ie, through identity adapter), which can be formulated as:
\begin{align}
&  \theta^{(t+1)} = \theta^{(t)} - M \odot \nabla_{\theta} \mathcal{L} \ , \\
&   M = Z \cdot M_m + (1 - Z) \cdot M_i \ .
\end{align}
The mask conditions are defined as:
\begin{align}
\begin{cases}
M_m[l] = 1 \iff M_m \cdot \theta[l] = \theta_m \ ,\\
M_i[k] = 1 \iff M_i \cdot \theta[k] = \theta_i \ ,
\end{cases}
\end{align}
where $\mathcal{L}$ denotes the video diffusion reconstruction loss. The adapter parameters $\theta$ are split into motion ($\theta_m$) and identity ($\theta_i$) components using binary masks $M_m$ and $M_i$.

Simultaneously, we keep a frozen adapter in a waiting state to inform modal expertise for active adapters during forward propagation, enabling two-dimensional features referencing within current data streams. The features from the frozen adapter act as intrinsic regularization to constrain dimension overfitting, thereby facilitating mutual reference learning without interference.

\subsection{StageBlender Controller}
\begin{figure}[t]
    \centering
    \includegraphics[width=\linewidth]{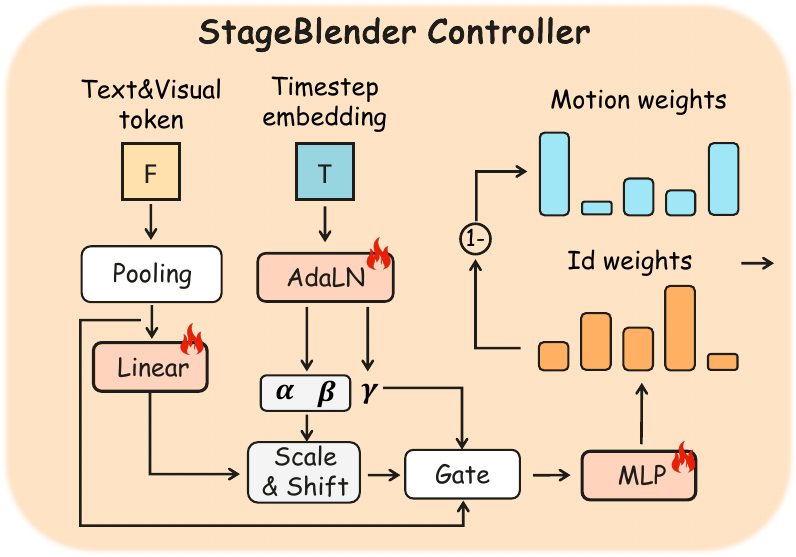}
    \caption{\textbf{Illustration of proposed StageBlender Controller}, which employs an Adaptive LayerNorm mechanism that modulates text-visual feature based on timestep-conditional embeddings, then maps the feature to multiple groups after residual gated connections. These scaled weights are subsequently routed to their respective DA-Blocks for processing.
    }
    \label{fig:hyper}
\end{figure}
\begin{figure*}[t]
    \centering
    \includegraphics[width=1\linewidth]{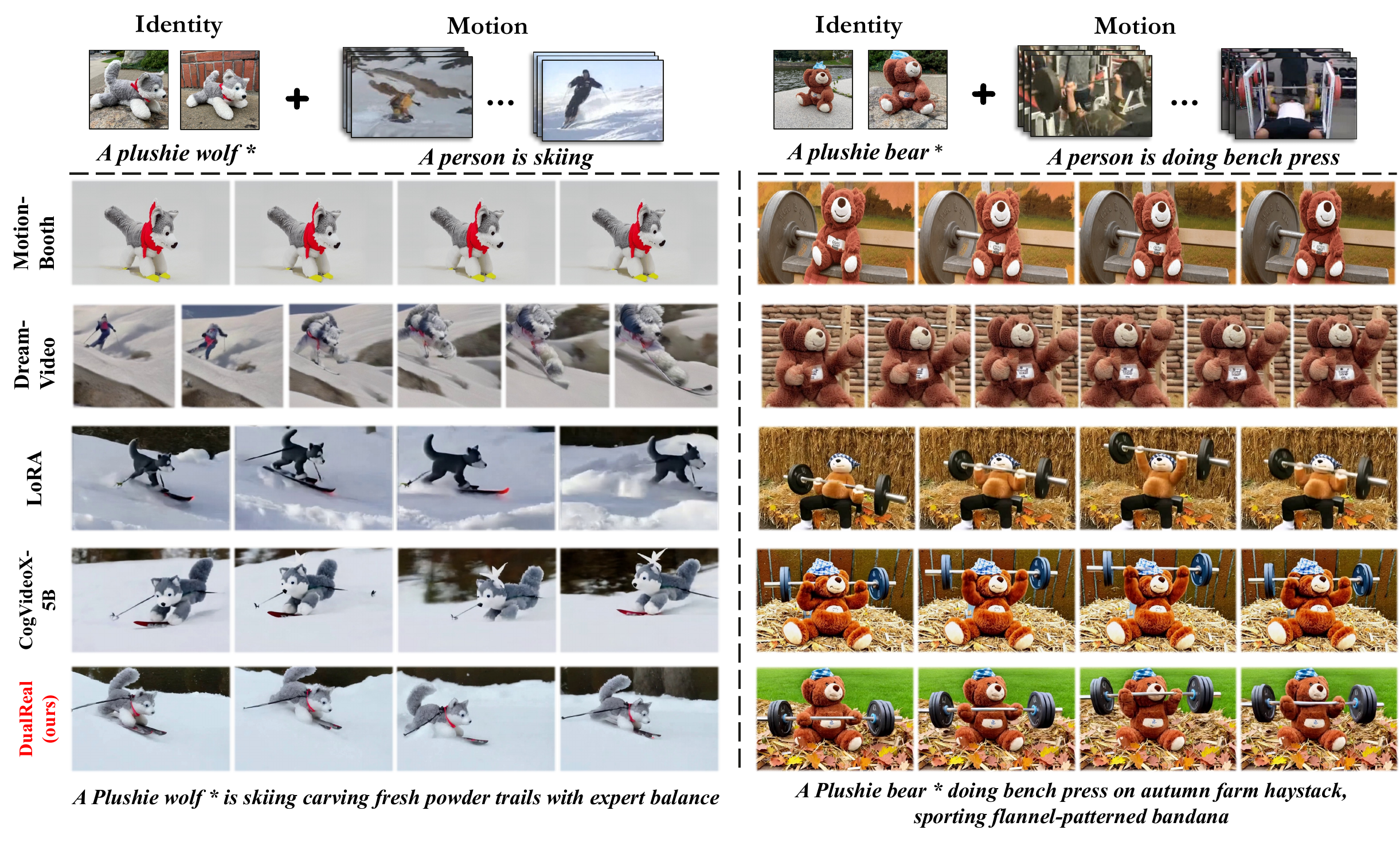}
    \caption{\textbf{Qualitative comparison with existing methods.} Compared with other methods, DualReal achieves high identity consistency with coherent motion, demonstrating the advantage of joint training in balancing pattern conflicts.}
    \label{fig:mainresult}
\end{figure*}
\label{sec:hypernet}
Furthermore, in order to resolve dimensional competition across the processing stage, we propose the \textit{StageBlender Controller} that governs dimensional contributions through time-aware conditioning of block-level scaling coefficients, which empowers the \textit{DA-Block} to adaptively allocate specific dimension shares (\ie, achieve granularity decoupling) through the mechanism detailed in \cref{fig:hyper}. Specifically, this module dynamically generates multiple sets of scaling weights according to denoising timestep embedding and the fused text-visual features. For the input feature $f_{\text{in}}=[f_\text{text}^1,f_\text{visual}^1] \in \mathbb{R}^{B \times (n_t+n_v) \times c}$, the processing flow first extracts salient features through pooling, then adaptively modulates them via DiT Adaptive LayerNorm ~\cite{yang2024cogvideox} with injected timestep embeddings $t$. This operation can be formulated as:
\begin{align}
&  f^{'}= \text{{Pooling}}(f_{\text{in}},\ \text{dim=1})*\mathbf{W}\ ,  \\
&  f^{''}=\text{MLP}\left(\text{LaynerNorm}(f')\right)*\mathbf{\alpha}+\mathbf{\beta} \ ,
\end{align}
where the $\mathbf{W}\in \mathbb{R}^{c \times t_{\text{dim}}}$ denotes the weight matrix with $t_{\text{dim}}$ as the channel dimension of timestep embedding, and $\alpha$, $\beta$ are defined as:
\begin{align}
& \mathbf{h} = \text{MLP}\left(\text{SiLU}(\mathbf{t})\right)\ ,\\
& \mathbf{\alpha},\ \mathbf{\beta},\ \mathbf{\gamma}=\mathbf{h}_{:d}, \mathbf{h}_{d:2d}, \mathbf{h}_{2d:3d}\ .
\end{align}
The computed weight coefficients are then integrated to enable gated fusion between the timestep and visual text tokens, as formulated below:
\begin{align}
& f_g = f^{''}+\mathbf{\gamma}*f^{'}.
\end{align}
Through empirical analysis of DiT-based denoising architectures, we observe that deeper blocks inherently specialize in processing concrete, fine-grained features. To enhance hierarchical decoupling, we implement a downward-propagating MLP that transforms integrated features into weight groups, as formalized below:
\begin{align}
 \underbrace{\mathbf{\omega}^{(1)},...,\mathbf{\omega}^{(n)}}_\text{Weights groups} = \ \text{softmax}\left(\mathbf{\Gamma}  \cdot(\text{MLP}(f_g)) \right)\  ,
\end{align}
where $\mathbf\Gamma$ is the projection operator: $\mathbb{R}^{L} \to \mathbb{R}^n$. The $L$ denotes the DiT block depth, and $n$ specifies the number of disentangled weight groups; each group then sequentially controls its assigned layers via parameter assignment.
\section{Experiment}
\begin{table*}[t]
    \centering
    \begin{tabular}{cccccccl}
        \toprule
        \multirow{2}{*}{\textbf{\small{Method}}} & 
        \multicolumn{1}{c}{\textbf{\textit{\small{Text Consistency}}}} & 
        \multicolumn{2}{c}{\textbf{\textit{\small{Identity Similarity}}}} & 
        \multicolumn{4}{c}{\textbf{\textit{\small{Motion Quality}}}} \\
        \cmidrule(lr){2-2} \cmidrule(lr){3-4} \cmidrule(lr){5-8} 
        & \textbf{\footnotesize{CLIP-T} $\uparrow$} & \textbf{\footnotesize{CLIP-I} $ \uparrow$} & \textbf{\footnotesize{DINO-I} $\uparrow$} & \textbf{\footnotesize{T.Flickering} $\uparrow$} & \textbf{\footnotesize{T.Cons} $\uparrow$} & \textbf{\footnotesize{Motion Smoothness} $\uparrow$} & \textbf{\footnotesize{Dynamic Degree}} \\
        \midrule
        \small{MotionBooth}~\cite{wu2024motionbooth} & 0.317 & \colorbox{myblue}{0.566} & \colorbox{myblue}{0.459} & \colorbox{myblue}{0.962} & 0.972 & \colorbox{myblue}{0.973} & \colorbox{mypink}{10.95\ ( -1.07 )}\\
        \small{LoRA}~\cite{hu2022lora} & \colorbox{myblue}{0.323} & 0.425 & 0.286 & 0.956 & \colorbox{myblue}{0.976} & \colorbox{myblue}{0.973} & 25.34\ ( +13.32 )  \\
        \footnotesize{CogVideoX-5B}~\cite{yang2024cogvideox} & \colorbox{mypink}{0.336} & 0.521 & 0.424 & 0.947 & 0.973 & 0.965 & 26.51 ( +14.49 ) \\
        \small{DreamVideo}~\cite{wei2024dreamvideo} & 0.278 & 0.458 & 0.334 & 0.949 & 0.963 & 0.968 & 8.841\ ( -3.18 ) \\
        \textbf{\small{DualReal (Ours)}} & \colorbox{myblue}{0.323} & \colorbox{mypink}{0.629} & \colorbox{mypink}{0.551} & \colorbox{mypink}{0.965} & \colorbox{mypink}{0.983} & \colorbox{mypink}{0.978} & \colorbox{myblue}{14.96\ ( +2.94 )} \\
        \bottomrule
    \end{tabular}
    \caption{
    \textbf{Quantitative comparison of personalization video generation for customized subject and motion.} We highlight the \colorbox{mypink}{best} and \colorbox{myblue}{second-best} values for each metric. ``T.Cons" and ``T.Flickering" denote Temporal Consistency and Temporal Flickering, respectively. Compared with other methods, \textit{DualReal} achieved average improvements of \textbf{21.7\%} on CLIP-I and \textbf{31.8\%} on DINO-I, recorded the best results on three motion quality metrics (T.Cons, Motion Smoothness, and Temporal Flickering), and ranked second on CLIP-T. The motion datasets achieve an average Dynamic Degree of \textbf{12.02}, and parenthetical values quantify the current method’s deviation from this benchmark to determine the intensity consistency of movement.
    }
    \label{tab:main}
\end{table*}
\subsection{Setup}
\noindent\textbf{Datasets.} The evaluation datasets are divided into two components: identity images and motion videos. For identity customization, 50 subjects are strictly selected from previous works~\cite{kumari2023multiconceptcustomizationtexttoimagediffusion, liu2023cones2customizableimage} and Internet collections (including pet, plush, \etc), with each subject containing 3–10 images. For motion customization, 21 motion sequences with challenging dynamic patterns are collected from public datasets \cite{soomro2012ucf101dataset101human, soomro2015action}. Additionally, each case is provided with 50 various prompts containing different editability (\ie, decoration or environment) to evaluate the method's editability and scene versatility sequentially.

\noindent\textbf{Baselines.} Among existing methods, DreamVideo \cite{wei2024dreamvideo} achieves customization of both identity and motion. For fair comparison, we implement two approaches from the same DiT backbone: (1) CogVideoX-5B~\cite{yang2024cogvideox}: Sequential full-parameter fine-tuning with identity then motion data as in the DreamBooth~\cite{ruiz2023dreambooth} paradigm. (2) LoRA fine-tuning~\cite{hu2022lora}: Separate training of two LoRA modules for identity and motion, then fuse their parameters during inference.
Additionally, the identity module of MotionBooth \cite{wu2024motionbooth} introduces irrelevant random videos during training to preserve the model’s motion capability, so we compare our approach with this method as well. In summary, we evaluate our results against DreamVideo, CogVideoX-5B, LoRA fine-tuning, and MotionBooth to provide a more comprehensive performance analysis.

\noindent\textbf{Evaluation metrics.}  We use seven metrics across three dimensions. (1) \textit{Text-Video Consistency} is measured by CLIP-T scores, computed as the CLIP~\cite{radford2021learning} cosine similarity between text prompts and all generated frames. (2) \textit{Identity Fidelity} is quantified using DINO-I and CLIP-I scores, which assess feature similarity between generated frames and reference identity images via DINO ViTS/16~\cite{caron2021emerging} and enhanced CLIP~\cite{guie2023winner} embeddings, respectively. (3) \textit{Temporal Motion Quality} is evaluated with four metrics: T-Cons~\cite{esser2023structure} for temporal consistency, Motion Smoothness (MS) for global fluidity, Temporal Flickering (TF) for high-frequency inconsistencies measured by mean absolute differences between adjacent frames, and Dynamic Degree (DD) leveraging RAFT optical flow estimation~\cite{teed2020raft} to quantify motion intensity (\textit{We quantify the method's deviation from the benchmark to determine the intensity consistency of movement}). Notably, MS, TF, and DD are adopted from the comprehensive video benchmark VBench~\cite{huang2024vbench}.

\subsection{Main Results}
\noindent\textbf{Qualitative results.} Qualitative experiments in \cref{fig:mainresult} show that while MotionBooth maintains identity fidelity, it fails to model motion patterns effectively. DreamVideo suffers from pattern conflicts during inference, resulting in inconsistent identity. Similarly, CogVideoX-5B and LoRA struggle to preserve identity due to their decoupled training methods. In contrast, DualReal achieves high identity consistency with coherent motion, demonstrating the advantage of joint training in balancing pattern conflicts.

\noindent\textbf{Quantitative results.}
As shown in \cref{tab:main}, \textit{DualReal} achieved average improvements of \textbf{21.7\%} on CLIP-I and \textbf{31.8\%} on DINO-I, recorded the best results on three motion quality metrics (T.Cons, Motion Smoothness, and Temporal Flickering), and ranked second on CLIP-T. Although our DD metric for quantifying motion intensity is not high, we evaluated all motion data and found an average DD of \textbf{\textit{12.02}}. Our metric deviates slightly from it, proving there is no collapse in motion amplitude. Overall, our method significantly enhances motion coherence and identity fidelity while preserving text consistency, further validating our adaptive joint training approach.
\subsection{Ablation Studies}
We evaluate our method by conducting ablation studies on a smaller evaluation subset, with the observed trends aligning with those of the main evaluation set. Additional ablation results are provided in the supplementary.
\begin{table} [t]
    \centering
    \small  
    \setlength{\tabcolsep}{1pt}  
    \begin{tabular}{ccccc}
        \toprule
        \textbf{Settings}  & \textbf{CLIP-T} & \textbf{CLIP-I} & \textbf{DINO-I}& \textbf{DD}\\
        \midrule
        w/o \small{Dual-aware Adaptation}   & 0.334 & 0.616 & 0.647 & 3.51(-5.53)\\
        w/o \small{StageBlender Controller} & \colorbox{mypink}{0.346} & 0.619 & 0.652 & 5.70(-3.31) \\
        w/o \small{Weight Groups} & \colorbox{myblue}{0.335} & \colorbox{myblue}{0.662} & \colorbox{myblue}{0.766} & \colorbox{myblue}{5.83(-3.12)} \\
        \textbf{ours} & 0.333 & \colorbox{mypink}{0.674} & \colorbox{mypink}{0.771} & \colorbox{mypink}{6.34(-2.70)}  \\
        \bottomrule
    \end{tabular}
    \caption{
    \textbf{Quantitative ablation studies} on each component. We implement Dual-aware Adaptation removal by separately training the two modalities and directly blending their parameters during inference, following the approach of DreamVideo. The motion datasets achieve an average Dynamic Degree of \textbf{9.04}.
    }
    \label{tab:ablation}
\end{table}
\begin{figure}[t]
    \centering
    \includegraphics[width=1\linewidth]{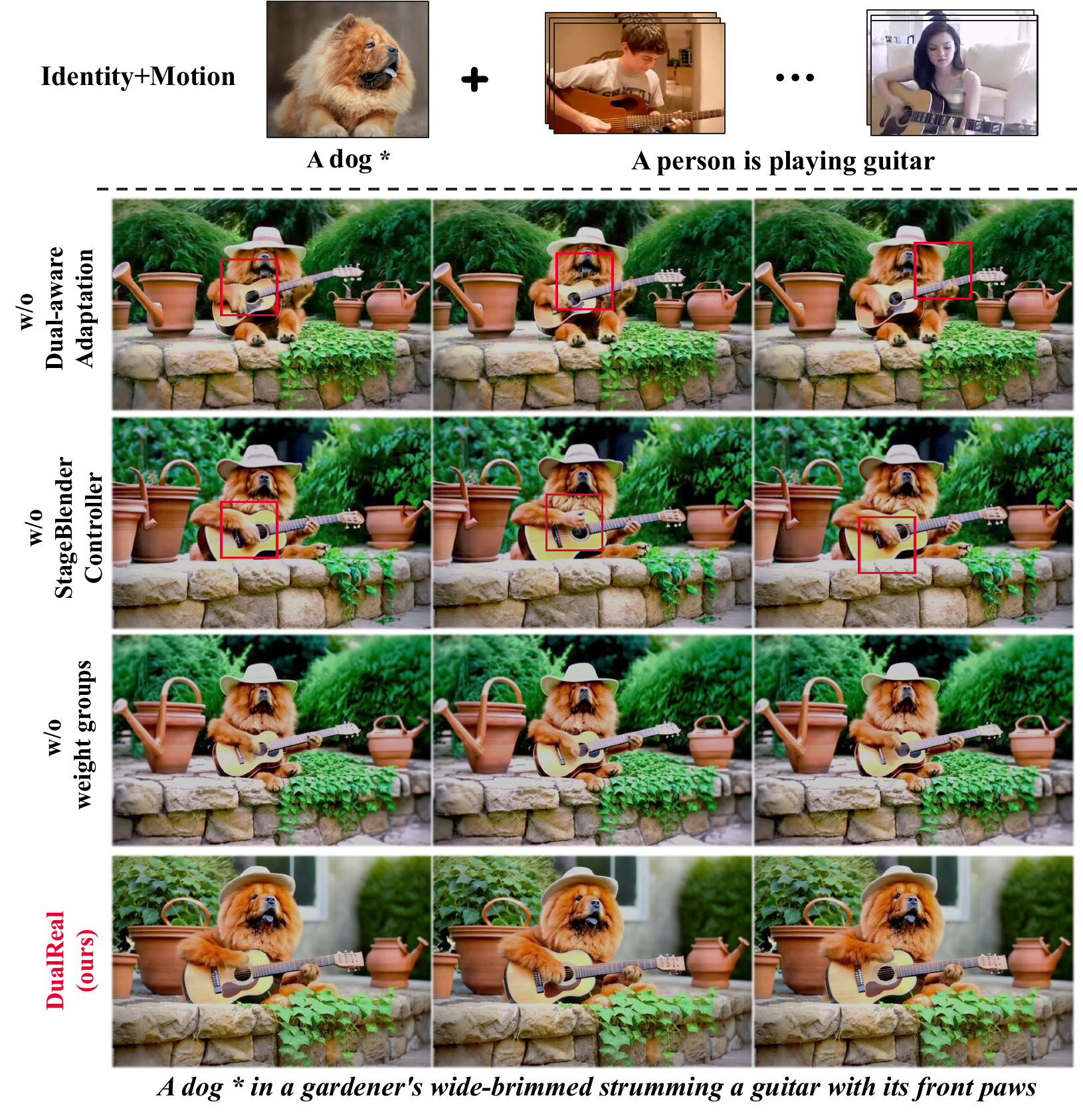}
    \caption{\textbf{Qualitative ablation studies on each component.}
    (1) Omitting Dual-aware Adaptation introduces \textbf{artifacts on the subject’s hands and chin}, significantly reducing clarity. Using fixed weights for the dimensional adapters without the StageBlender Controller \textit{causes the hands to become overly adapted to the motion pattern}, and removing weight grouping \textit{reduces identity fidelity and background detail}.
    }
    \label{fig:ablation}
\end{figure}

\begin{table} [t]
    \centering
    \small  
    \setlength{\tabcolsep}{3.5pt}  
    \begin{tabular}{ccccc}
        \toprule
        \textbf{Group Cardinality}  & \textbf{CLIP-T} & \textbf{CLIP-I} & \textbf{DINO-I}& \textbf{DD}  \\
        \midrule
        n=1 & 0.335 & \colorbox{myblue}{0.662} & \colorbox{myblue}{0.766} & 5.83\ (-3.21) \\
        n=2  & \colorbox{mypink}{0.343} & 0.632 & 0.660 & 5.49\ (-3.55)  \\
        n=42  & \colorbox{myblue}{0.336} & 0.631 & 0.706 & \colorbox{myblue}{6.24\ (-2.80)}  \\
        \textbf{ours}(n=7) & 0.333 & \colorbox{mypink}{0.674} & \colorbox{mypink}{0.771} & \colorbox{mypink}{6.34\ (-2.70)}  \\
        \bottomrule
    \end{tabular}
    \caption{\textbf{Quantitative ablation studies of group cardinality.} The results suggest that very small groups may lack sufficient context and overly large groups may dilute crucial details, making \textbf{balanced group cardinality essential for optimal performance.}}
    \label{tab:ablation-2}
\end{table}

\noindent\textbf{Quantitative experiment.}
The ablation study in \cref{tab:ablation} shows that removing \textit{Dual-aware Adaptation} or the \textit{StageBlender Controller} slightly increases the text consistency metric CLIP-T but significantly decreases identity similarity and motion intensity, highlighting the need for joint dimension training and granular control. Additionally, metric changes diminish slightly when weight groups are removed.

\noindent\textbf{Qualitative experiment.} Qualitative results in \cref{fig:ablation} reveal that omitting \textit{Dual-aware Adaptation} produces artifacts on the hands and chin, degrading clarity.
Using fixed weights for the dimensional adapters without the \textit{StageBlender Controller} (\ie, direct fusion at inference) overfits the hands to motion patterns. Removing weight grouping (\ie, uniform block modulation) weakens identity fidelity and background detail. These observations confirm that every component is essential for high‑quality customized generation.

\noindent\textbf{Effectiveness of group cardinality.} As shown in \cref{tab:ablation-2}, CLIP-I and DINO-I performance declines when group cardinality is either very small or very large, while a balanced group size (n=7) yields the best results. This suggests that very small groups may lack sufficient context and overly large groups may dilute crucial details, making balanced group cardinality essential for optimal performance.

\subsection{Visual analysis of StageBlender Controller}
\begin{figure}[t]
    \centering
    \includegraphics[width=1\linewidth]{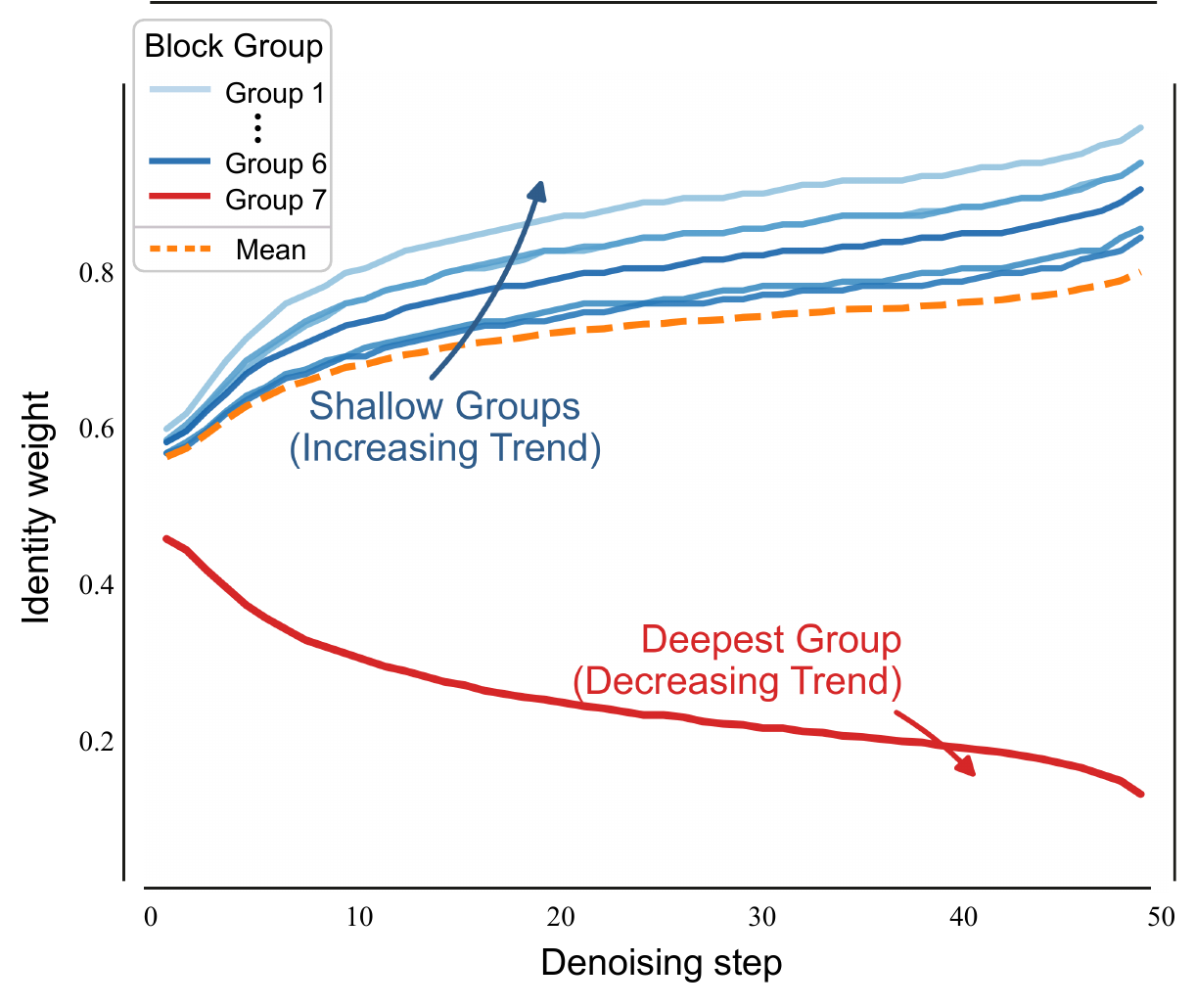}
    \caption{\textbf{Controller Visual Analysis.} We show the Identity Weights trends across denoising steps for different block depths.
    (1) As denoising progresses, the diffusion model’s emphasis shifts monotonically between identity and motion, \textit{with a growing focus on identity(orange dashed line)};
    (2) The \textit{deepest block group} exhibits an inverse pattern, \ie, with the denoising process \textit{increasingly prioritizing motion coherence} modeling.
    }
    \label{fig:visual}
\end{figure}

As \cref{fig:visual} shows, shallow blocks (Groups 1–6, blue) progressively increase identity weights during denoising, emphasizing early identity preservation. In contrast, the deepest block (Group 7, red) steadily raises motion weights to enhance motion modeling. Overall (orange dashed line), as denoising advances, the model increasingly prioritizes identity preservation over motion generation, highlighting the distinct roles of different network depths.
These observations further confirm that: (1) as denoising progresses, the diffusion model's emphasis shifts monotonically between identity and motion, with \textbf{a growing focus on identity}; (2) DiT networks of different depths divide the tasks of modeling identity and motion differently at each denoising step, with \textbf{the deepest network focusing on motion patterns} and showing increased enhancement as denoising advances.

\section{Conclusion}
In this paper, we propose \textit{DualReal}, a novel approach for customized video generation given a subject and motion. \textit{DualReal} adaptively trains identity and motion jointly, resolving dimensional conflicts and enabling universal sample customization. Our framework leverages the prior from one dimension to guide the training of the other, while preventing information leakage through a regularization strategy. Simultaneously, we use a controller to guide the high-fidelity fusion of modes based on various denoising stages and DiT depths. Evaluated on a more comprehensive evaluation benchmark,
our method improves CLIP-I and DINO-I metrics by \textbf{21.7\%} and \textbf{31.8\%}, and achieves top performance on nearly all motion quality metrics, demonstrating the efficiency of our adaptive joint training framework.
\section{Acknowledgment}
This research is supported by Artificial Intelligence National Science and Technology Major Project 2023ZD0121200, and National Natural Science Foundation of China under Grant 62222212 and 623B2094.
{
    \small
    \bibliographystyle{ieeenat_fullname}
    \bibliography{main}

\begin{thebibliography}{59}
\providecommand{\natexlab}[1]{#1}
\providecommand{\url}[1]{\texttt{#1}}
\expandafter\ifx\csname urlstyle\endcsname\relax
  \providecommand{\doi}[1]{doi: #1}\else
  \providecommand{\doi}{doi: \begingroup \urlstyle{rm}\Url}\fi

\bibitem[An et~al.(2023)An, Zhang, Yang, Gupta, Huang, Luo, and Yin]{an2023latent}
Jie An, Songyang Zhang, Harry Yang, Sonal Gupta, Jia-Bin Huang, Jiebo Luo, and Xi Yin.
\newblock Latent-shift: Latent diffusion with temporal shift for efficient text-to-video generation.
\newblock \emph{arXiv preprint arXiv:2304.08477}, 2023.

\bibitem[Bai et~al.(2024)Bai, He, Wang, Guo, Hu, Liu, and Bian]{bai2024uniedit}
Jianhong Bai, Tianyu He, Yuchi Wang, Junliang Guo, Haoji Hu, Zuozhu Liu, and Jiang Bian.
\newblock Uniedit: A unified tuning-free framework for video motion and appearance editing.
\newblock \emph{arXiv preprint arXiv:2402.13185}, 2024.

\bibitem[Blattmann et~al.(2023)Blattmann, Dockhorn, Kulal, Mendelevitch, Kilian, Lorenz, Levi, English, Voleti, Letts, et~al.]{blattmann2023stable}
Andreas Blattmann, Tim Dockhorn, Sumith Kulal, Daniel Mendelevitch, Maciej Kilian, Dominik Lorenz, Yam Levi, Zion English, Vikram Voleti, Adam Letts, et~al.
\newblock Stable video diffusion: Scaling latent video diffusion models to large datasets.
\newblock \emph{arXiv preprint arXiv:2311.15127}, 2023.

\bibitem[Caron et~al.(2021)Caron, Touvron, Misra, J{\'e}gou, Mairal, Bojanowski, and Joulin]{caron2021emerging}
Mathilde Caron, Hugo Touvron, Ishan Misra, Herv{\'e} J{\'e}gou, Julien Mairal, Piotr Bojanowski, and Armand Joulin.
\newblock Emerging properties in self-supervised vision transformers.
\newblock In \emph{Proceedings of the IEEE/CVF international conference on computer vision}, pages 9650--9660, 2021.

\bibitem[Chefer et~al.(2024)Chefer, Zada, Paiss, Ephrat, Tov, Rubinstein, Wolf, Dekel, Michaeli, and Mosseri]{chefer2024still}
Hila Chefer, Shiran Zada, Roni Paiss, Ariel Ephrat, Omer Tov, Michael Rubinstein, Lior Wolf, Tali Dekel, Tomer Michaeli, and Inbar Mosseri.
\newblock Still-moving: Customized video generation without customized video data.
\newblock \emph{ACM Transactions on Graphics (TOG)}, 43\penalty0 (6):\penalty0 1--11, 2024.

\bibitem[Chen et~al.(2023{\natexlab{a}})Chen, Zhang, Wang, Duan, Zhou, and Zhu]{chen2023disenbooth}
Hong Chen, Yipeng Zhang, Xin Wang, Xuguang Duan, Yuwei Zhou, and Wenwu Zhu.
\newblock Disenbooth: Disentangled parameter-efficient tuning for subject-driven text-to-image generation.
\newblock \emph{arXiv preprint arXiv:2305.03374}, 3\penalty0 (4), 2023{\natexlab{a}}.

\bibitem[Chen et~al.(2025)Chen, Huang, Chen, Zheng, Zhang, and Mao]{chen2025customcontrast}
Nan Chen, Mengqi Huang, Zhuowei Chen, Yang Zheng, Lei Zhang, and Zhendong Mao.
\newblock Customcontrast: A multilevel contrastive perspective for subject-driven text-to-image customization.
\newblock In \emph{Proceedings of the AAAI Conference on Artificial Intelligence}, pages 2123--2131, 2025.

\bibitem[Chen et~al.(2023{\natexlab{b}})Chen, Hu, Li, Ruiz, Jia, Chang, and Cohen]{chen2023subject}
Wenhu Chen, Hexiang Hu, Yandong Li, Nataniel Ruiz, Xuhui Jia, Ming-Wei Chang, and William~W Cohen.
\newblock Subject-driven text-to-image generation via apprenticeship learning.
\newblock \emph{Advances in Neural Information Processing Systems}, 36:\penalty0 30286--30305, 2023{\natexlab{b}}.

\bibitem[Esser et~al.(2023)Esser, Chiu, Atighehchian, Granskog, and Germanidis]{esser2023structure}
Patrick Esser, Johnathan Chiu, Parmida Atighehchian, Jonathan Granskog, and Anastasis Germanidis.
\newblock Structure and content-guided video synthesis with diffusion models.
\newblock In \emph{Proceedings of the IEEE/CVF international conference on computer vision}, pages 7346--7356, 2023.

\bibitem[Gal et~al.(2022)Gal, Alaluf, Atzmon, Patashnik, Bermano, Chechik, and Cohen-Or]{gal2022image}
Rinon Gal, Yuval Alaluf, Yuval Atzmon, Or Patashnik, Amit~H Bermano, Gal Chechik, and Daniel Cohen-Or.
\newblock An image is worth one word: Personalizing text-to-image generation using textual inversion.
\newblock \emph{arXiv preprint arXiv:2208.01618}, 2022.

\bibitem[Geyer et~al.(2023)Geyer, Bar-Tal, Bagon, and Dekel]{geyer2023tokenflow}
Michal Geyer, Omer Bar-Tal, Shai Bagon, and Tali Dekel.
\newblock Tokenflow: Consistent diffusion features for consistent video editing.
\newblock \emph{arXiv preprint arXiv:2307.10373}, 2023.

\bibitem[Guo et~al.(2023)Guo, Yang, Rao, Liang, Wang, Qiao, Agrawala, Lin, and Dai]{guo2023animatediff}
Yuwei Guo, Ceyuan Yang, Anyi Rao, Zhengyang Liang, Yaohui Wang, Yu Qiao, Maneesh Agrawala, Dahua Lin, and Bo Dai.
\newblock Animatediff: Animate your personalized text-to-image diffusion models without specific tuning.
\newblock \emph{arXiv preprint arXiv:2307.04725}, 2023.

\bibitem[Han et~al.(2023)Han, Li, Zhang, Milanfar, Metaxas, and Yang]{han2023svdiff}
Ligong Han, Yinxiao Li, Han Zhang, Peyman Milanfar, Dimitris Metaxas, and Feng Yang.
\newblock Svdiff: Compact parameter space for diffusion fine-tuning.
\newblock In \emph{Proceedings of the IEEE/CVF International Conference on Computer Vision}, pages 7323--7334, 2023.

\bibitem[He et~al.(2024)He, Liu, Qian, Wang, Hu, Cao, Yan, and Zhang]{he2024id}
Xuanhua He, Quande Liu, Shengju Qian, Xin Wang, Tao Hu, Ke Cao, Keyu Yan, and Jie Zhang.
\newblock Id-animator: Zero-shot identity-preserving human video generation.
\newblock \emph{arXiv preprint arXiv:2404.15275}, 2024.

\bibitem[Hendrycks and Gimpel(2016)]{hendrycks2016gaussian}
Dan Hendrycks and Kevin Gimpel.
\newblock Gaussian error linear units (gelus).
\newblock \emph{arXiv preprint arXiv:1606.08415}, 2016.

\bibitem[Ho et~al.(2022)Ho, Salimans, Gritsenko, Chan, Norouzi, and Fleet]{ho2022video}
Jonathan Ho, Tim Salimans, Alexey Gritsenko, William Chan, Mohammad Norouzi, and David~J Fleet.
\newblock Video diffusion models.
\newblock \emph{Advances in Neural Information Processing Systems}, 35:\penalty0 8633--8646, 2022.

\bibitem[Hong et~al.(2022)Hong, Ding, Zheng, Liu, and Tang]{hong2022cogvideo}
Wenyi Hong, Ming Ding, Wendi Zheng, Xinghan Liu, and Jie Tang.
\newblock Cogvideo: Large-scale pretraining for text-to-video generation via transformers.
\newblock \emph{arXiv preprint arXiv:2205.15868}, 2022.

\bibitem[Hu et~al.(2022)Hu, Shen, Wallis, Allen-Zhu, Li, Wang, Wang, Chen, et~al.]{hu2022lora}
Edward~J Hu, Yelong Shen, Phillip Wallis, Zeyuan Allen-Zhu, Yuanzhi Li, Shean Wang, Lu Wang, Weizhu Chen, et~al.
\newblock Lora: Low-rank adaptation of large language models.
\newblock \emph{ICLR}, 1\penalty0 (2):\penalty0 3, 2022.

\bibitem[Huang et~al.(2022)Huang, Mao, Wang, Wang, and Zhang]{huang2022dse}
Mengqi Huang, Zhendong Mao, Penghui Wang, Quan Wang, and Yongdong Zhang.
\newblock Dse-gan: Dynamic semantic evolution generative adversarial network for text-to-image generation.
\newblock In \emph{Proceedings of the 30th ACM International Conference on Multimedia}, pages 4345--4354, 2022.

\bibitem[Huang et~al.(2023)Huang, Mao, Chen, and Zhang]{huang2023towards}
Mengqi Huang, Zhendong Mao, Zhuowei Chen, and Yongdong Zhang.
\newblock Towards accurate image coding: Improved autoregressive image generation with dynamic vector quantization.
\newblock In \emph{Proceedings of the IEEE/CVF Conference on Computer Vision and Pattern Recognition}, pages 22596--22605, 2023.

\bibitem[Huang et~al.(2024{\natexlab{a}})Huang, Mao, Liu, He, and Zhang]{huang2024realcustom}
Mengqi Huang, Zhendong Mao, Mingcong Liu, Qian He, and Yongdong Zhang.
\newblock Realcustom: narrowing real text word for real-time open-domain text-to-image customization.
\newblock In \emph{Proceedings of the IEEE/CVF Conference on Computer Vision and Pattern Recognition}, pages 7476--7485, 2024{\natexlab{a}}.

\bibitem[Huang et~al.(2024{\natexlab{b}})Huang, He, Yu, Zhang, Si, Jiang, Zhang, Wu, Jin, Chanpaisit, et~al.]{huang2024vbench}
Ziqi Huang, Yinan He, Jiashuo Yu, Fan Zhang, Chenyang Si, Yuming Jiang, Yuanhan Zhang, Tianxing Wu, Qingyang Jin, Nattapol Chanpaisit, et~al.
\newblock Vbench: Comprehensive benchmark suite for video generative models.
\newblock In \emph{Proceedings of the IEEE/CVF Conference on Computer Vision and Pattern Recognition}, pages 21807--21818, 2024{\natexlab{b}}.

\bibitem[Jabri et~al.(2022)Jabri, Fleet, and Chen]{jabri2022scalable}
Allan Jabri, David Fleet, and Ting Chen.
\newblock Scalable adaptive computation for iterative generation.
\newblock \emph{arXiv preprint arXiv:2212.11972}, 2022.

\bibitem[Jiang et~al.(2024)Jiang, Wu, Yang, Si, Lin, Qiao, Loy, and Liu]{jiang2024videobooth}
Yuming Jiang, Tianxing Wu, Shuai Yang, Chenyang Si, Dahua Lin, Yu Qiao, Chen~Change Loy, and Ziwei Liu.
\newblock Videobooth: Diffusion-based video generation with image prompts.
\newblock In \emph{Proceedings of the IEEE/CVF Conference on Computer Vision and Pattern Recognition}, pages 6689--6700, 2024.

\bibitem[Kong et~al.(2024)Kong, Tian, Zhang, Min, Dai, Zhou, Xiong, Li, Wu, Zhang, et~al.]{kong2024hunyuanvideo}
Weijie Kong, Qi Tian, Zijian Zhang, Rox Min, Zuozhuo Dai, Jin Zhou, Jiangfeng Xiong, Xin Li, Bo Wu, Jianwei Zhang, et~al.
\newblock Hunyuanvideo: A systematic framework for large video generative models.
\newblock \emph{arXiv preprint arXiv:2412.03603}, 2024.

\bibitem[Kumari et~al.(2023)Kumari, Zhang, Zhang, Shechtman, and Zhu]{kumari2023multiconceptcustomizationtexttoimagediffusion}
Nupur Kumari, Bingliang Zhang, Richard Zhang, Eli Shechtman, and Jun-Yan Zhu.
\newblock Multi-concept customization of text-to-image diffusion, 2023.

\bibitem[Liu et~al.(2023)Liu, Zhang, Shen, Zheng, Zhu, Feng, Liu, Zhao, Zhou, and Cao]{liu2023cones2customizableimage}
Zhiheng Liu, Yifei Zhang, Yujun Shen, Kecheng Zheng, Kai Zhu, Ruili Feng, Yu Liu, Deli Zhao, Jingren Zhou, and Yang Cao.
\newblock Cones 2: Customizable image synthesis with multiple subjects, 2023.

\bibitem[Ma et~al.(2024)Ma, Zhou, Yeh, Wang, Li, Yang, Dong, Keutzer, and Feng]{ma2024magic}
Ze Ma, Daquan Zhou, Chun-Hsiao Yeh, Xue-She Wang, Xiuyu Li, Huanrui Yang, Zhen Dong, Kurt Keutzer, and Jiashi Feng.
\newblock Magic-me: Identity-specific video customized diffusion.
\newblock \emph{arXiv preprint arXiv:2402.09368}, 2024.

\bibitem[Mao et~al.(2024)Mao, Huang, Ding, Liu, He, and Zhang]{mao2024realcustom++}
Zhendong Mao, Mengqi Huang, Fei Ding, Mingcong Liu, Qian He, and Yongdong Zhang.
\newblock Realcustom++: Representing images as real-word for real-time customization.
\newblock \emph{arXiv preprint arXiv:2408.09744}, 2024.

\bibitem[Menapace et~al.(2024)Menapace, Siarohin, Skorokhodov, Deyneka, Chen, Kag, Fang, Stoliar, Ricci, Ren, et~al.]{menapace2024snap}
Willi Menapace, Aliaksandr Siarohin, Ivan Skorokhodov, Ekaterina Deyneka, Tsai-Shien Chen, Anil Kag, Yuwei Fang, Aleksei Stoliar, Elisa Ricci, Jian Ren, et~al.
\newblock Snap video: Scaled spatiotemporal transformers for text-to-video synthesis.
\newblock In \emph{Proceedings of the IEEE/CVF Conference on Computer Vision and Pattern Recognition}, pages 7038--7048, 2024.

\bibitem[Molad et~al.(2023)Molad, Horwitz, Valevski, Acha, Matias, Pritch, Leviathan, and Hoshen]{molad2023dreamix}
Eyal Molad, Eliahu Horwitz, Dani Valevski, Alex~Rav Acha, Yossi Matias, Yael Pritch, Yaniv Leviathan, and Yedid Hoshen.
\newblock Dreamix: Video diffusion models are general video editors.
\newblock \emph{arXiv preprint arXiv:2302.01329}, 2023.

\bibitem[OpenAI(2024)]{sora2024}
OpenAI.
\newblock Sora, 2024.

\bibitem[Peebles and Xie(2023)]{peebles2023scalable}
William Peebles and Saining Xie.
\newblock Scalable diffusion models with transformers.
\newblock In \emph{Proceedings of the IEEE/CVF international conference on computer vision}, pages 4195--4205, 2023.

\bibitem[Qing et~al.(2024)Qing, Zhang, Wang, Wang, Wei, Zhang, Gao, and Sang]{qing2024hierarchical}
Zhiwu Qing, Shiwei Zhang, Jiayu Wang, Xiang Wang, Yujie Wei, Yingya Zhang, Changxin Gao, and Nong Sang.
\newblock Hierarchical spatio-temporal decoupling for text-to-video generation.
\newblock In \emph{Proceedings of the IEEE/CVF Conference on Computer Vision and Pattern Recognition}, pages 6635--6645, 2024.

\bibitem[Radford et~al.(2021)Radford, Kim, Hallacy, Ramesh, Goh, Agarwal, Sastry, Askell, Mishkin, Clark, et~al.]{radford2021learning}
Alec Radford, Jong~Wook Kim, Chris Hallacy, Aditya Ramesh, Gabriel Goh, Sandhini Agarwal, Girish Sastry, Amanda Askell, Pamela Mishkin, Jack Clark, et~al.
\newblock Learning transferable visual models from natural language supervision.
\newblock In \emph{International conference on machine learning}, pages 8748--8763. PmLR, 2021.

\bibitem[Ren et~al.(2024)Ren, Zhou, Yang, Shi, Liu, Liu, Kwon, and Shrivastava]{ren2024customize}
Yixuan Ren, Yang Zhou, Jimei Yang, Jing Shi, Difan Liu, Feng Liu, Mingi Kwon, and Abhinav Shrivastava.
\newblock Customize-a-video: One-shot motion customization of text-to-video diffusion models.
\newblock In \emph{European Conference on Computer Vision}, pages 332--349. Springer, 2024.

\bibitem[Ronneberger et~al.(2015)Ronneberger, Fischer, and Brox]{ronneberger2015u}
Olaf Ronneberger, Philipp Fischer, and Thomas Brox.
\newblock U-net: Convolutional networks for biomedical image segmentation.
\newblock In \emph{Medical image computing and computer-assisted intervention--MICCAI 2015: 18th international conference, Munich, Germany, October 5-9, 2015, proceedings, part III 18}, pages 234--241. Springer, 2015.

\bibitem[Ruiz et~al.(2023)Ruiz, Li, Jampani, Pritch, Rubinstein, and Aberman]{ruiz2023dreambooth}
Nataniel Ruiz, Yuanzhen Li, Varun Jampani, Yael Pritch, Michael Rubinstein, and Kfir Aberman.
\newblock Dreambooth: Fine tuning text-to-image diffusion models for subject-driven generation.
\newblock In \emph{Proceedings of the IEEE/CVF conference on computer vision and pattern recognition}, pages 22500--22510, 2023.

\bibitem[Ruiz et~al.(2024)Ruiz, Li, Jampani, Wei, Hou, Pritch, Wadhwa, Rubinstein, and Aberman]{ruiz2024hyperdreambooth}
Nataniel Ruiz, Yuanzhen Li, Varun Jampani, Wei Wei, Tingbo Hou, Yael Pritch, Neal Wadhwa, Michael Rubinstein, and Kfir Aberman.
\newblock Hyperdreambooth: Hypernetworks for fast personalization of text-to-image models.
\newblock In \emph{Proceedings of the IEEE/CVF conference on computer vision and pattern recognition}, pages 6527--6536, 2024.

\bibitem[Shao et~al.(2023)Shao, Yu, Zhao, and Ge]{guie2023winner}
Shihao Shao, Lijun Yu, Yifan Zhao, and Yixiao Ge.
\newblock 1st place solution in google universal image embedding challenge.
\newblock https://github.com/ShihaoShao-GH/1st-Place-Solution-in-Google-Universal-Image-Embedding, 2023.

\bibitem[Shi et~al.(2024)Shi, Xiong, Lin, and Jung]{shi2024instantbooth}
Jing Shi, Wei Xiong, Zhe Lin, and Hyun~Joon Jung.
\newblock Instantbooth: Personalized text-to-image generation without test-time finetuning.
\newblock In \emph{Proceedings of the IEEE/CVF conference on computer vision and pattern recognition}, pages 8543--8552, 2024.

\bibitem[Singer et~al.(2022)Singer, Polyak, Hayes, Yin, An, Zhang, Hu, Yang, Ashual, Gafni, et~al.]{singer2022make}
Uriel Singer, Adam Polyak, Thomas Hayes, Xi Yin, Jie An, Songyang Zhang, Qiyuan Hu, Harry Yang, Oron Ashual, Oran Gafni, et~al.
\newblock Make-a-video: Text-to-video generation without text-video data.
\newblock \emph{arXiv preprint arXiv:2209.14792}, 2022.

\bibitem[Smith et~al.(2023)Smith, Hsu, Zhang, Hua, Kira, Shen, and Jin]{smith2023continual}
James~Seale Smith, Yen-Chang Hsu, Lingyu Zhang, Ting Hua, Zsolt Kira, Yilin Shen, and Hongxia Jin.
\newblock Continual diffusion: Continual customization of text-to-image diffusion with c-lora.
\newblock \emph{arXiv preprint arXiv:2304.06027}, 2023.

\bibitem[Soomro and Zamir(2015)]{soomro2015action}
Khurram Soomro and Amir~R Zamir.
\newblock Action recognition in realistic sports videos.
\newblock In \emph{Computer vision in sports}, pages 181--208. Springer, 2015.

\bibitem[Soomro et~al.(2012)Soomro, Zamir, and Shah]{soomro2012ucf101dataset101human}
Khurram Soomro, Amir~Roshan Zamir, and Mubarak Shah.
\newblock Ucf101: A dataset of 101 human actions classes from videos in the wild, 2012.

\bibitem[Tan et~al.(2024)Tan, Gong, Wang, Zhang, Zheng, Zheng, Zheng, Chen, and Yang]{tan2024animate}
Shuai Tan, Biao Gong, Xiang Wang, Shiwei Zhang, Dandan Zheng, Ruobing Zheng, Kecheng Zheng, Jingdong Chen, and Ming Yang.
\newblock Animate-x: Universal character image animation with enhanced motion representation.
\newblock \emph{arXiv preprint arXiv:2410.10306}, 2024.

\bibitem[Teed and Deng(2020)]{teed2020raft}
Zachary Teed and Jia Deng.
\newblock Raft: Recurrent all-pairs field transforms for optical flow.
\newblock In \emph{Computer Vision--ECCV 2020: 16th European Conference, Glasgow, UK, August 23--28, 2020, Proceedings, Part II 16}, pages 402--419. Springer, 2020.

\bibitem[Wang et~al.(2023{\natexlab{a}})Wang, Yuan, Chen, Zhang, Wang, and Zhang]{wang2023modelscope}
Jiuniu Wang, Hangjie Yuan, Dayou Chen, Yingya Zhang, Xiang Wang, and Shiwei Zhang.
\newblock Modelscope text-to-video technical report.
\newblock \emph{arXiv preprint arXiv:2308.06571}, 2023{\natexlab{a}}.

\bibitem[Wang et~al.(2023{\natexlab{b}})Wang, Yuan, Zhang, Chen, Wang, Zhang, Shen, Zhao, and Zhou]{wang2023videocomposer}
Xiang Wang, Hangjie Yuan, Shiwei Zhang, Dayou Chen, Jiuniu Wang, Yingya Zhang, Yujun Shen, Deli Zhao, and Jingren Zhou.
\newblock Videocomposer: Compositional video synthesis with motion controllability.
\newblock \emph{Advances in Neural Information Processing Systems}, 36:\penalty0 7594--7611, 2023{\natexlab{b}}.

\bibitem[Wang et~al.(2024)Wang, Li, Zhu, Guo, Dou, and Li]{wang2024customvideo}
Zhao Wang, Aoxue Li, Lingting Zhu, Yong Guo, Qi Dou, and Zhenguo Li.
\newblock Customvideo: Customizing text-to-video generation with multiple subjects.
\newblock \emph{arXiv preprint arXiv:2401.09962}, 2024.

\bibitem[Wei et~al.(2023)Wei, Zhang, Ji, Bai, Zhang, and Zuo]{wei2023elite}
Yuxiang Wei, Yabo Zhang, Zhilong Ji, Jinfeng Bai, Lei Zhang, and Wangmeng Zuo.
\newblock Elite: Encoding visual concepts into textual embeddings for customized text-to-image generation.
\newblock In \emph{Proceedings of the IEEE/CVF International Conference on Computer Vision}, pages 15943--15953, 2023.

\bibitem[Wei et~al.(2024)Wei, Zhang, Qing, Yuan, Liu, Liu, Zhang, Zhou, and Shan]{wei2024dreamvideo}
Yujie Wei, Shiwei Zhang, Zhiwu Qing, Hangjie Yuan, Zhiheng Liu, Yu Liu, Yingya Zhang, Jingren Zhou, and Hongming Shan.
\newblock Dreamvideo: Composing your dream videos with customized subject and motion.
\newblock In \emph{Proceedings of the IEEE/CVF Conference on Computer Vision and Pattern Recognition}, pages 6537--6549, 2024.

\bibitem[Wu et~al.(2024{\natexlab{a}})Wu, Li, Zeng, Zhang, Zhou, Li, Tong, and Chen]{wu2024motionbooth}
Jianzong Wu, Xiangtai Li, Yanhong Zeng, Jiangning Zhang, Qianyu Zhou, Yining Li, Yunhai Tong, and Kai Chen.
\newblock Motionbooth: Motion-aware customized text-to-video generation.
\newblock \emph{arXiv preprint arXiv:2406.17758}, 2024{\natexlab{a}}.

\bibitem[Wu et~al.(2023)Wu, Ge, Wang, Lei, Gu, Shi, Hsu, Shan, Qie, and Shou]{wu2023tune}
Jay~Zhangjie Wu, Yixiao Ge, Xintao Wang, Stan~Weixian Lei, Yuchao Gu, Yufei Shi, Wynne Hsu, Ying Shan, Xiaohu Qie, and Mike~Zheng Shou.
\newblock Tune-a-video: One-shot tuning of image diffusion models for text-to-video generation.
\newblock In \emph{Proceedings of the IEEE/CVF International Conference on Computer Vision}, pages 7623--7633, 2023.

\bibitem[Wu et~al.(2024{\natexlab{b}})Wu, Zhang, Wang, Zhou, Zheng, Qi, Shan, and Li]{wu2024customcrafter}
Tao Wu, Yong Zhang, Xintao Wang, Xianpan Zhou, Guangcong Zheng, Zhongang Qi, Ying Shan, and Xi Li.
\newblock Customcrafter: Customized video generation with preserving motion and concept composition abilities.
\newblock \emph{arXiv preprint arXiv:2408.13239}, 2024{\natexlab{b}}.

\bibitem[Yang et~al.(2024)Yang, Teng, Zheng, Ding, Huang, Xu, Yang, Hong, Zhang, Feng, et~al.]{yang2024cogvideox}
Zhuoyi Yang, Jiayan Teng, Wendi Zheng, Ming Ding, Shiyu Huang, Jiazheng Xu, Yuanming Yang, Wenyi Hong, Xiaohan Zhang, Guanyu Feng, et~al.
\newblock Cogvideox: Text-to-video diffusion models with an expert transformer.
\newblock \emph{arXiv preprint arXiv:2408.06072}, 2024.

\bibitem[Zhang et~al.(2024)Zhang, Wu, Liu, Zhao, Ran, Gu, Gao, and Shou]{zhang2024show}
David~Junhao Zhang, Jay~Zhangjie Wu, Jia-Wei Liu, Rui Zhao, Lingmin Ran, Yuchao Gu, Difei Gao, and Mike~Zheng Shou.
\newblock Show-1: Marrying pixel and latent diffusion models for text-to-video generation.
\newblock \emph{International Journal of Computer Vision}, pages 1--15, 2024.

\bibitem[Zhao et~al.(2024)Zhao, Gu, Wu, Zhang, Liu, Wu, Keppo, and Shou]{zhao2024motiondirector}
Rui Zhao, Yuchao Gu, Jay~Zhangjie Wu, David~Junhao Zhang, Jia-Wei Liu, Weijia Wu, Jussi Keppo, and Mike~Zheng Shou.
\newblock Motiondirector: Motion customization of text-to-video diffusion models.
\newblock In \emph{European Conference on Computer Vision}, pages 273--290. Springer, 2024.

\bibitem[Zhou et~al.(2022)Zhou, Wang, Yan, Lv, Zhu, and Feng]{zhou2022magicvideo}
Daquan Zhou, Weimin Wang, Hanshu Yan, Weiwei Lv, Yizhe Zhu, and Jiashi Feng.
\newblock Magicvideo: Efficient video generation with latent diffusion models.
\newblock \emph{arXiv preprint arXiv:2211.11018}, 2022.

\end{thebibliography}
}
\clearpage
\begin{figure*}[t]
    \centering
    \includegraphics[width=0.9\linewidth]{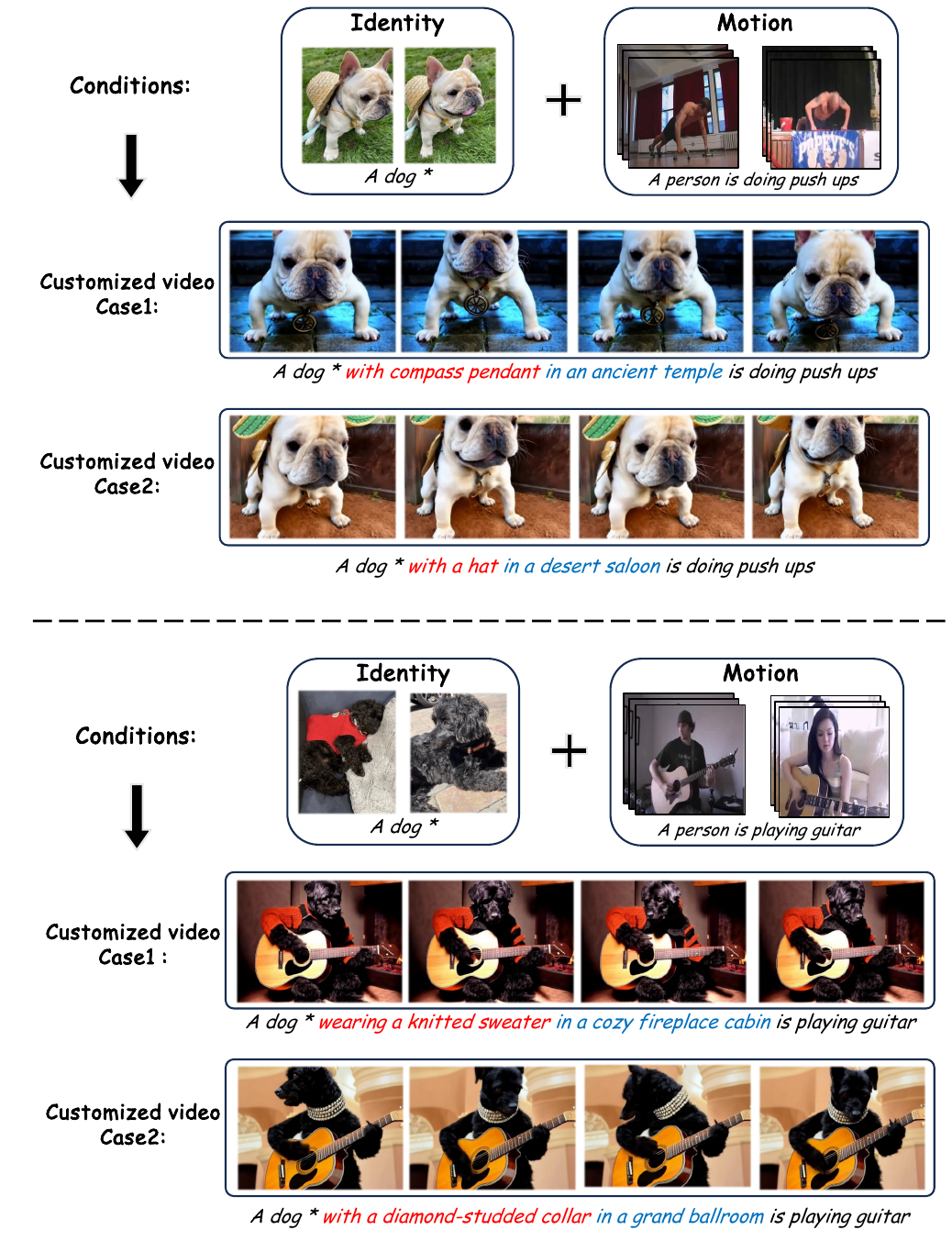}
    \caption{
    Generated customization results of our proposed novel paradigm \textbf{\textit{DualReal}}. Given subject images and motion videos, \textit{DualReal} generates high-quality customized identity and motion simultaneously, without compromising the consistency of either dimension.
    }
    \label{fig:sup1}
\end{figure*}

\begin{figure*}[t]
    \centering
    \includegraphics[width=0.9\linewidth]{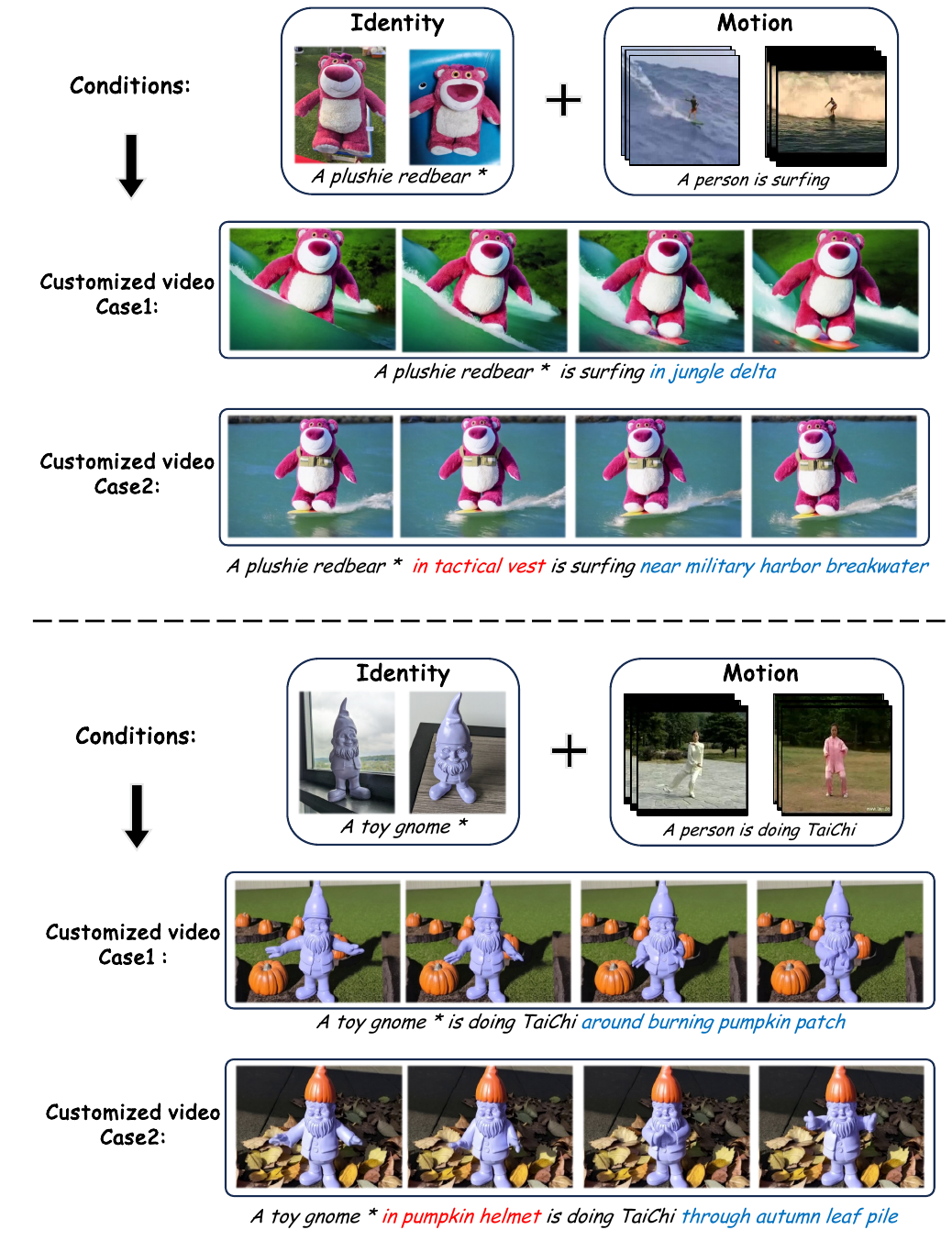}
    \caption{
    Generated customization results of our proposed novel paradigm \textbf{\textit{DualReal}}. Given subject images and motion videos, \textit{DualReal} generates high-quality customized identity and motion simultaneously, without compromising the consistency of either dimension.
    }
    \label{fig:sup2}
\end{figure*}

\begin{figure*}[t]
    \centering
    \includegraphics[width=0.95\linewidth]{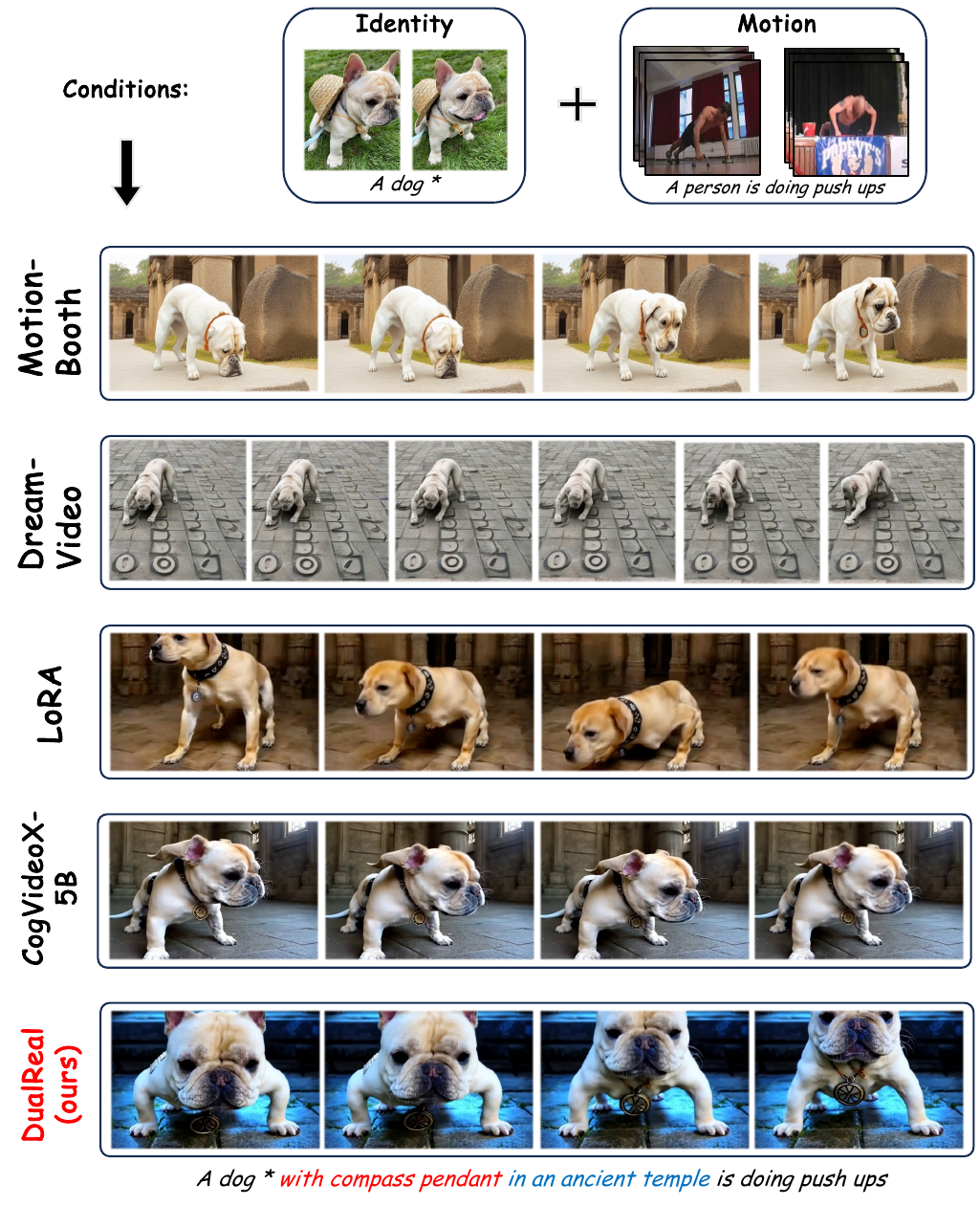}
    \caption{\textbf{More Qualitative comparison with existing methods.} The result shows that while MotionBooth maintains identity fidelity, it fails to model motion patterns effectively. DreamVideo suffers from pattern conflicts during inference, resulting in inconsistent identity. Similarly, CogVideoX-5B and LoRA struggle to preserve identity due to their decoupled training methods. In contrast, DualReal achieves high identity consistency with coherent motion, demonstrating the advantage of joint training in balancing pattern conflicts.}
    \label{fig:mainresult_appendix}
\end{figure*}
\begin{figure*}[t]
    \centering
    \includegraphics[width=1\linewidth]{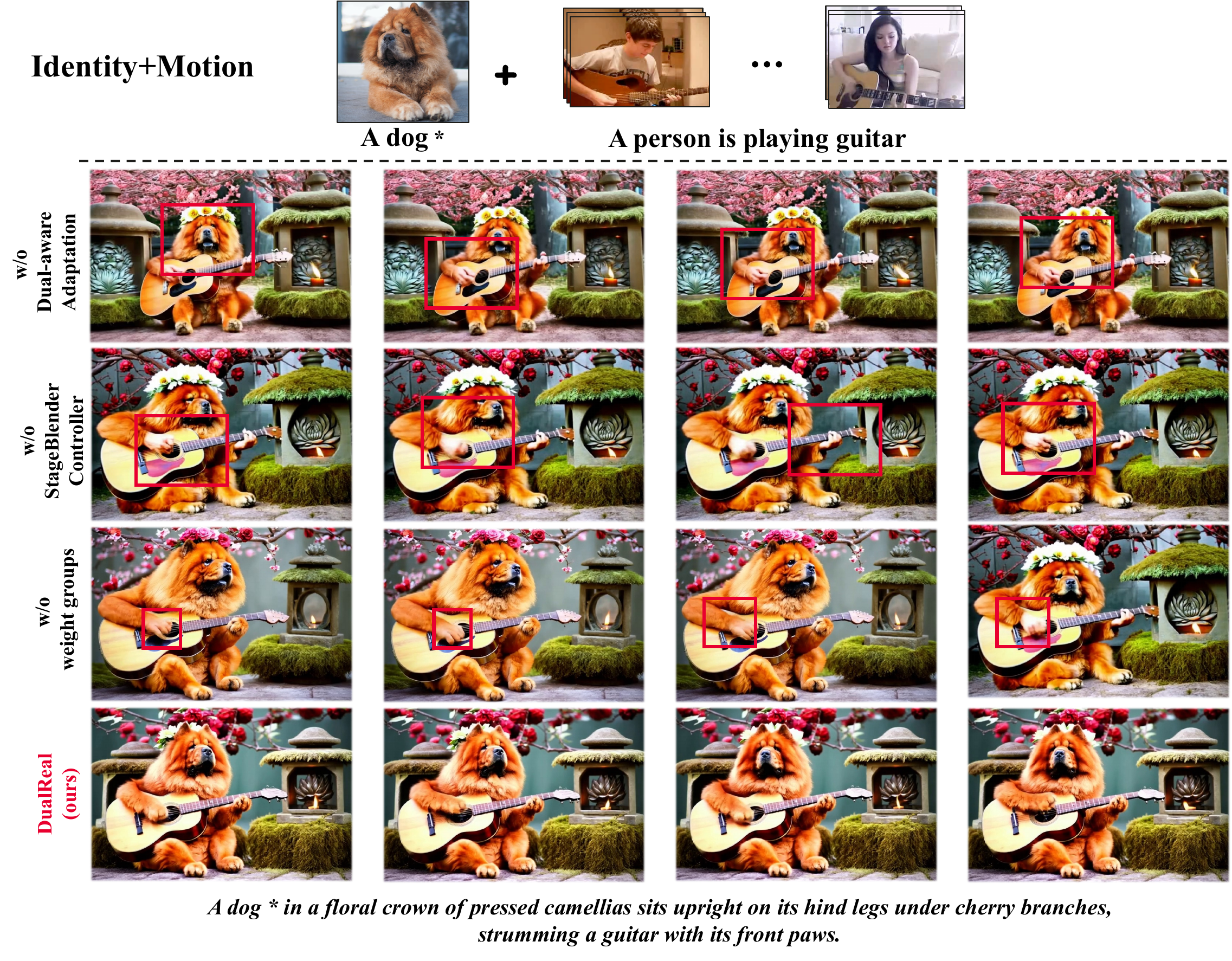}
    \caption{
    \textbf{Qualitative ablations studies on each component.}
    Omitting Dual-aware Adaptation introduces artifacts on the subject’s hands, significantly reducing clarity. Moreover, using fixed weights for the dimensional adapters without the StageBlender Controller causes over-adaptation to the motion pattern, and omitting weight grouping further undermines identity fidelity.
    }
    \label{fig:ablation_appendix}
\end{figure*}

\section{Supplementary}
\subsection{Experimental Details}
This section describes the implementation of our primary experiments and ablation studies. For each method, we provide detailed information on the setup. We list hyperparameter values, data pre‑processing and post‑processing steps, training schedules, and evaluation protocols. All information is provided to ensure reproducibility and clarity.

\noindent\textbf{DualReal.}
We run 1,000 training steps for every test case. We set $\gamma=0.5$ so that each step has a 50\% chance of motion training. The learning rate is 1e-3. We use the AdamW optimizer to ensure stable convergence and effective weight regularization. Under these settings, our method consistently produces high-quality customized videos. Each output contains 49 frames at a resolution of 480×720 pixels.

\noindent\textbf{Baseline.}
For MotionBooth, we adopt LaVie‑base as the text‑to‑video backbone, set the learning rate to 5e-6, train for 300 steps with a batch size of 10 using the unique token “sks” and the AdamW optimizer.
For both LoRA and full‑parameter fine‑tuning, we follow the official CogVideoX training code: LoRA uses a learning rate of 1e-3 with 300 identity steps and 300 motion steps, while full fine‑tuning uses a learning rate of 1e-4 with 200 identity steps and 130 motion steps.
For DreamVideo, we build on the ModelScopeT2V V1.5 base model and follow the recommended schedule, first training the identity stage for 3000 steps (batch size 4, learning rate 1e-4), then continuing identity training for 500 steps (batch size 4, learning rate 1e-5), and finally running multi‑video motion training for 600 steps (batch size 2, learning rate 1e-5).

\noindent\textbf{Prompts.}
Given a target identity and motion, we employ a large language model to enrich the prompt by appending details, such as clothing styles, accessories, and situating the subject in diverse settings that align with the intended action. This automated prompt expansion introduces both semantic variety and environmental complexity, enabling us to rigorously evaluate the extent to which our customized video framework can accurately interpret and render nuanced textual edits.

\subsection{More Main Results}
To highlight the differences among methods, we conduct a comprehensive qualitative comparison between \textit{DualReal} and several state‑of‑the‑art baselines. Whereas prior approaches often sacrifice either identity fidelity or motion realism, \textit{DualReal} delivers both: it preserves distinctive identity features while producing smooth, temporally consistent motion.
This dual achievement stems directly from our joint training scheme, which aligns identity and motion objectives within a unified optimization process and thereby resolves the inherent conflicts between static appearance and dynamic behavior. The result is a harmonious fusion of identity and motion, as shown in \cref{fig:mainresult_appendix}.

\subsection{More Ablation Results}
We provide additional qualitative results in \cref{fig:ablation_appendix} that further validate the influence of each component, aligning with the descriptions provided in the main paper. Omitting Dual‑aware Adaptation introduces visible artifacts, especially around the hands and chin, that markedly degrade clarity. Replacing our StageBlender Controller with direct fusion (i.e., using fixed adapter weights at inference) causes the model to over-adapt to motion dynamics. Eliminating weight grouping so that all blocks receive uniform modulation leads to weakened identity preservation and a loss of background detail. Together, these findings demonstrate that every module in our pipeline is critical for achieving high‑quality, customized video generation.

\subsection{More Cases}
The \textit{DualReal} framework dynamically tailors its dual processing pathways to any combination of user-supplied identity references and motion sequences, irrespective of their complexity tier. By automatically calibrating to input difficulty—from simple to intricate actions—it synthesizes personalized 720x480 resolution videos comprising 49 temporally consistent frames. Crucially, the system rigorously preserves subject identity characteristics while ensuring smooth motion transitions across all generated content. This dual-path adaptability addresses the core challenge of reconciling visual authenticity with kinematic continuity, establishing a generalized solution for user-customized video generation across diverse input scenarios. We further demonstrate the generation effect of our method on different cases and prompts as shown in \cref{fig:sup1} and \cref{fig:sup2}.

\end{document}